\documentclass{article}
\usepackage[utf8]{inputenc}
\usepackage{amssymb}
\usepackage{amsmath}
\usepackage{tabularx}
\usepackage{hyperref}
\usepackage{fancybox}
\usepackage{tikz}
\usepackage{authblk}
\usepackage{physics}

\DeclareMathOperator*{\argmin}{arg\,min}

\author[1,*]{Nabeel Sarwar}
\author[2,*]{Wilson Gregory}
\author[2,*]{George A Kevrekidis} 

\author[2,3,$^\dagger$]{Soledad Villar}

\author[4,$^\dagger$]{\hspace{2cm}
 Bianca Dumitrascu}

\affil[1]{Center for Data Science, New York University, NY 10012, USA}
\affil[2]{Department of Applied Mathematics and Statistics, Johns Hopkins University, MD 21218, USA}
\affil[3]{Mathematical Institute for Data Science, Johns Hopkins University, MD 21218 USA}
\affil[4]{Department of Computer Science and Technology, University of Cambridge, UK}

\date{}
\usepackage{graphicx}
\usepackage[numbers]{natbib}

\usepackage[margin=0.6in]{geometry}

\title{MarkerMap: nonlinear marker selection for single-cell studies}

\usepackage{natbib}
\usepackage{graphicx}

\usepackage{float}
\usepackage{subcaption}

\usepackage{amsmath}
\usepackage{bm}

\begin{document}
\parindent 0pt
\parskip 0.2cm
\maketitle
\textit{\footnotesize{
$*$ these authors contributed equally to this work}}
\textit{\footnotesize{
$^\dagger$ joint correspondence to: bmd39@cam.ac.uk, soledad.villar@jhu.edu}}
\begin{abstract}

Single-cell RNA-seq data allow the quantification of cell type differences across a growing set of biological contexts. However, pinpointing a small subset of genomic features explaining this variability can be ill-defined and computationally intractable. Here we introduce MarkerMap, a generative model for selecting minimal gene sets which are maximally informative of cell type origin and enable whole transcriptome reconstruction. MarkerMap provides a scalable framework for both supervised marker selection, aimed at identifying specific cell type populations, and unsupervised marker selection, aimed at gene expression imputation and reconstruction. We benchmark MarkerMap’s competitive performance against previously published approaches on real single cell gene expression data sets. MarkerMap is available as a pip installable package, as a community resource aimed at developing explainable machine learning techniques for enhancing interpretability in single-cell studies.

%publicly available  \url{https://github.com/Computational-Morphogenomics-Group/MarkerMap}

\end{abstract}

\section*{Introduction}

Recent advances in genomics and microscopy enable the collection of single cell gene expression data (scRNA-seq) across cells from spatial \cite{lohoff2020highly} and temporal \cite{sladitschek2020morphoseq} coordinates. Understanding how cells aggregate information across spatio-temporal scales and how, in turn, gene expression variability reflects this aggregation process remains challenging. A particular experimental design challenge is due
to the fact that existing techniques (e.g., smFish \cite{codeluppi2018spatial}, seqFish 
\cite{lubeck2014single}, MERFISH \cite{chen2015spatially}, ISS \cite{ke2013situ}) rely on the pre-selection of a small number of target genes or \emph{markers}, incapable of capturing the full transcriptomic information required to characterize subtle differences in cell populations. Selecting the best such markers (\emph{marker selection}) is often statistically and computationally challenging, often a function of the nonlinearity of the data and the type of differences to be captured.

% what is the problem? and what are existing works

Marker selection is the product of both prior knowledge and computational analysis  of previously collected scRNA-seq data. In a nutshell, it is a dimensionality reduction task which enables downstream analysis such as visualization, cell type recovery or gene panel design for interventional studies. Akin to principal component analysis (PCA) \cite{hotelling1933analysis} or variational autoencoders (VAE) \cite{kingma2013auto}, both popular in the analysis of single-cell RNA-seq \cite{townes2019feature,svensson2020interpretable}, marker selection methods seek to describe cells as datapoints in a space of few coordinates. To this end, PCA and VAE based methodologies associate cells with a smaller set of latent coordinates representing aggregates of weighted groups of gene expression. In contrast, marker selection approaches seek interpretable representations, where coordinates represent genes directly, rather than linear or nonlinear combinations of genes.

% what are the solutions proposed in genomics?

Many methods have been proposed to select markers that best differentiate between a set of discrete, pre-defined cell type classes \cite{finak2015mast,delaney2019combinatorial, ibrahim2019genesorter, dumitrascu2021optimal, vargo2020rank, nelson2021smash}. These fall into two broad categories -- one-vs-all and gene panel methods. One-vs-all methods are most 
common \cite{finak2015mast,delaney2019combinatorial, ibrahim2019genesorter} and seek to determine, for each cell type, a set of genes that are differentially expressed in that \emph{one} cell type alone, when compared with \emph{all} the other cell types. In particular, RankCorr \cite{vargo2020rank}, a sparse selection approach inspired by the success of a related proteomic application \cite{conrad2017sparse}, offers theoretical guarantees and excellent experimental performance. Another recent algorithm with good performance, SMaSH \cite{nelson2021smash}, uses a neural network framework leveraging techniques from the interpretable machine learning literature \cite{shrikumar2017learning}. In contrast, gene panel methods seek to identify groups of genetic markers that jointly distinguish across cell types. ScGeneFit \cite{dumitrascu2021optimal}, for instance, employs linear programming to select markers that preserve the classification structure of the data, without identifying genes with individual cell types, and possibly selecting fewer genes as a result. It was defined as a linear programming relaxation of compressive classification, which asks for a projection to a low dimensional subspace where points with different labels remain separated \cite{mcwhirter2019squeezefit}. One-vs-all and gene panel alike, these methods are supervised:  they rely on a ground truth classification structure of the cells. Few unsupervised techniques exist -- SCMER \cite{liang2021single} is, to the best of our knowledge, the only genetic marker selection approach proposed that avoids explicit clustering by using nonlinear dimensionality reduction (UMAP) and manifold learning. Recent reviews on feature selection in genomics applications \cite{yang2021feature,pullin2022comparison} compare and contrast these marker selection methodologies in supervised, linear contexts. 

Further, solutions have been proposed to address the feature selection problem in non-genomic contexts as well. In linear settings, these include the popular $\ell_1$ regularization or Lasso  \cite{tibshirani1996regression}), and CUR decomposition \cite{mahoney2009cur}, while in nonlinear regression settings, outcomes are often predicted with neural networks \cite{lemhadri2021lassonet}. In language models, explainable deep learning algorithms have been developed to predict and interpret outcomes like review ratings or interview outcomes from texts where few significant words get highlighted as explanations for the outcome \cite{maddison_concrete, xie2019reparameterizable,abid2019concrete-ae, jang2016categorical}. 

%this paragraph is about our contributions;

In this paper, we introduce MarkerMap, a scalable and generative framework for nonlinear marker selection. Our objectives are two-fold: a) to provide a general method allowing for joint marker selection and full transcriptome reconstruction, b) to compare and contrast tools across different communities -- computational biology and explainable machine learning -- within a single, accessible computational framework centered around transcriptomic studies. As a result, MarkerMap exhibits several key 
features. First, MarkerMap scales to large data sets without the need for ad-hoc gene pruning. 

Second, it provides a joint setting for both supervised and unsupervised learning.
Third, it is generative, allowing for imputation to whole transcriptome levels from a reduced, informative number of markers. We provide a set of metrics to evaluate the quality of the imputations and compare the distributions of original transcriptomes with their reconstructions. Forth, its supervised option robustly tolerates small rates of labelling misclassification, which could emerge from processing and cell type assignment errors. We apply MarkerMap to real data, including cord blood mononuclear cells (CBMCs) assayed with different technologies,  longitudinal samples from mouse embryogenesis, and a developmental mouse brain single cell gene expression resource. Finally, a strong link exists between marker selection and the wider explainable machine learning literature \cite{abid2019concrete-ae, xie2019reparameterizable}. As both communities are rapidly evolving, there is an increasing need to systematically compare new and existing methods, with the goal of understanding their strengths and limitations. To address this need, we benchmark MarkerMap against existing marker selection approaches and related methodologies from the wider explainable machine learning literature. We make MakerMap available as a pip installable package. 

\begin{figure*}[htp]
    \centering
    \begin{subfigure}{\textwidth}
        \centering
        \includegraphics[width=\textwidth]{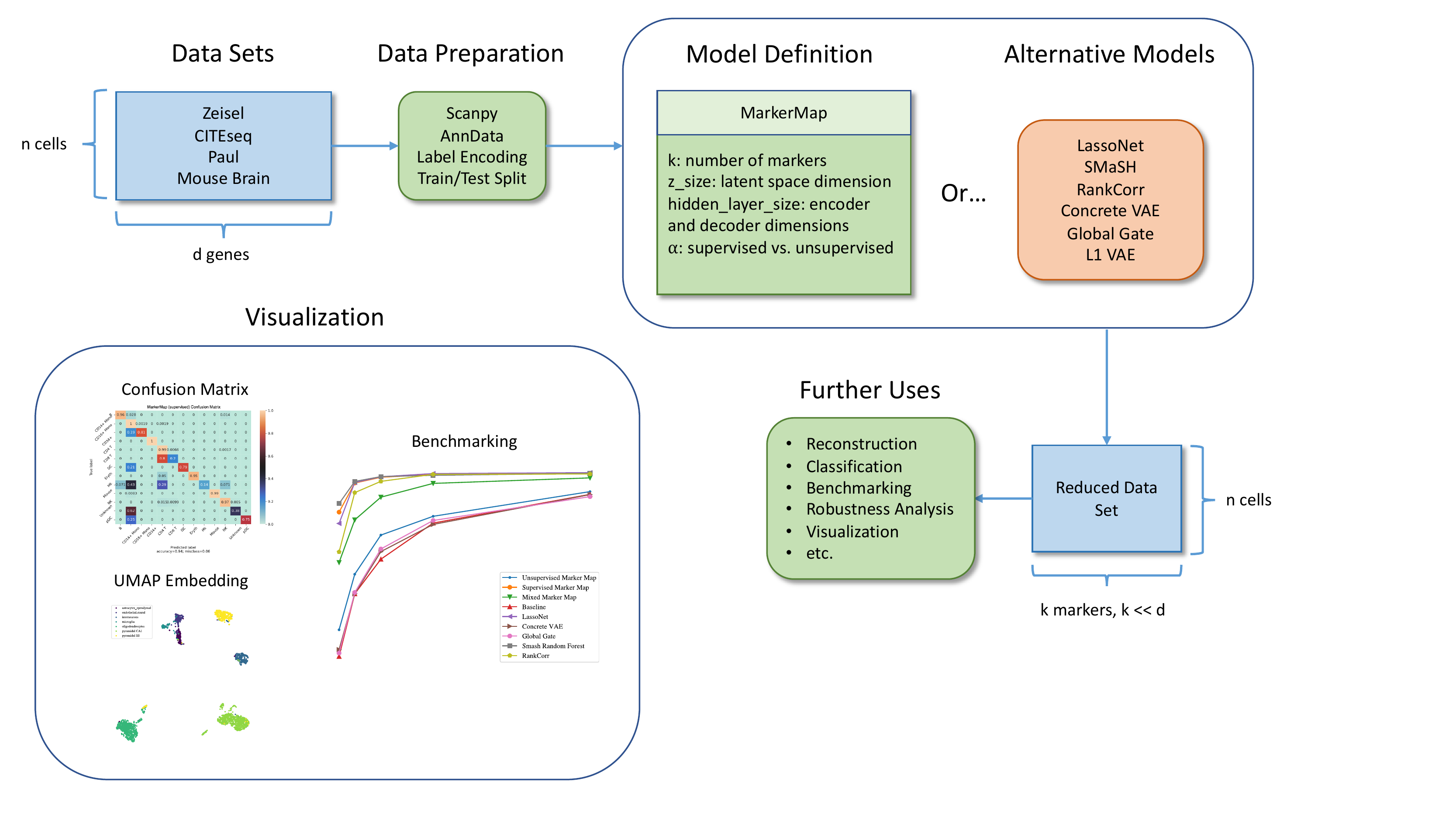}
        \caption{}
        \label{fig:markermap_pipeline}
    \end{subfigure}
    \begin{subfigure}{0.49\textwidth}
        \centering
        \includegraphics[width=\textwidth]{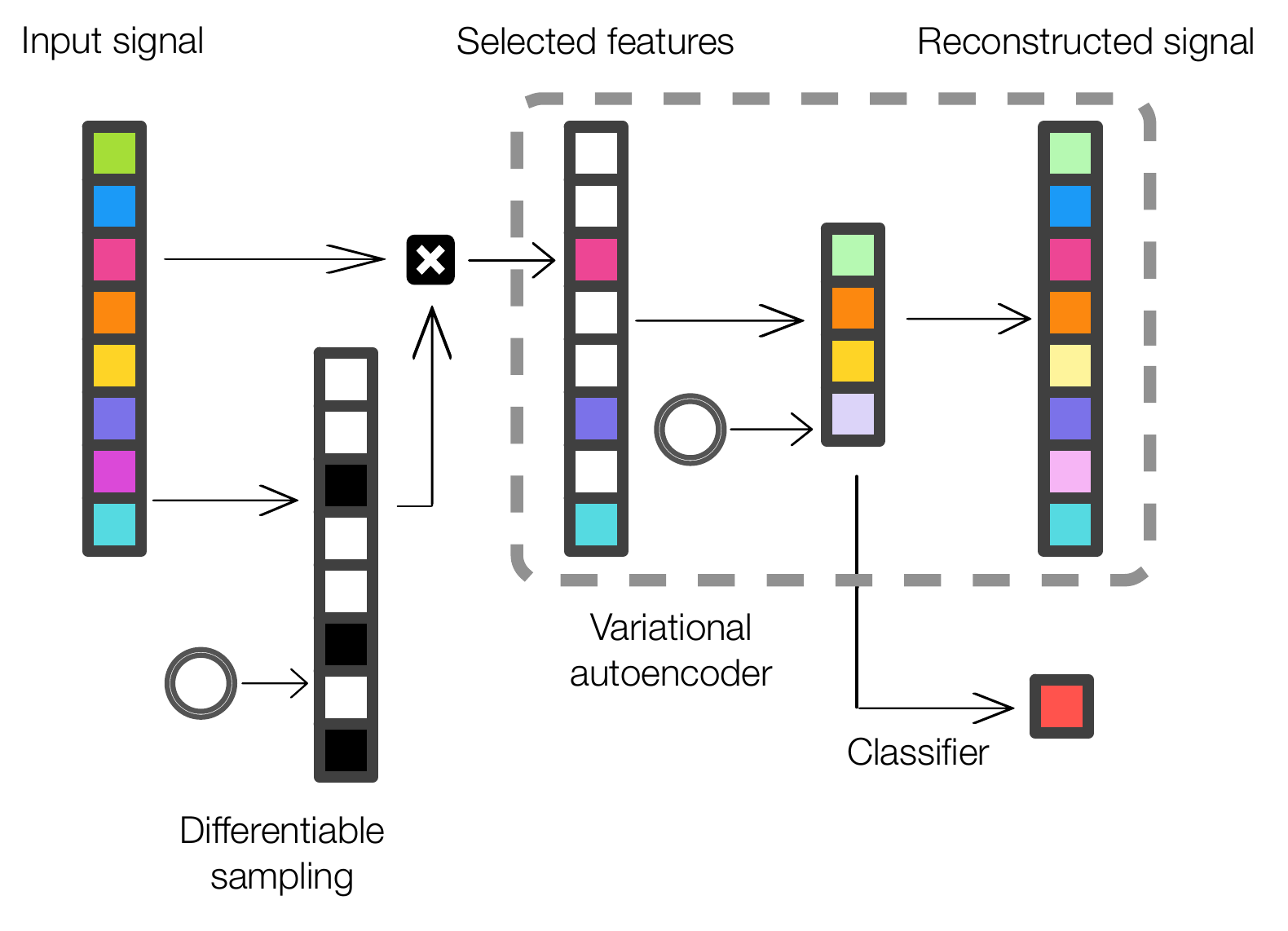}
        \caption{}
        \label{fig:architecture}
    \end{subfigure}
    \hfill
    \begin{subfigure}{0.49\textwidth}
        \centering
        \includegraphics[width=\textwidth]{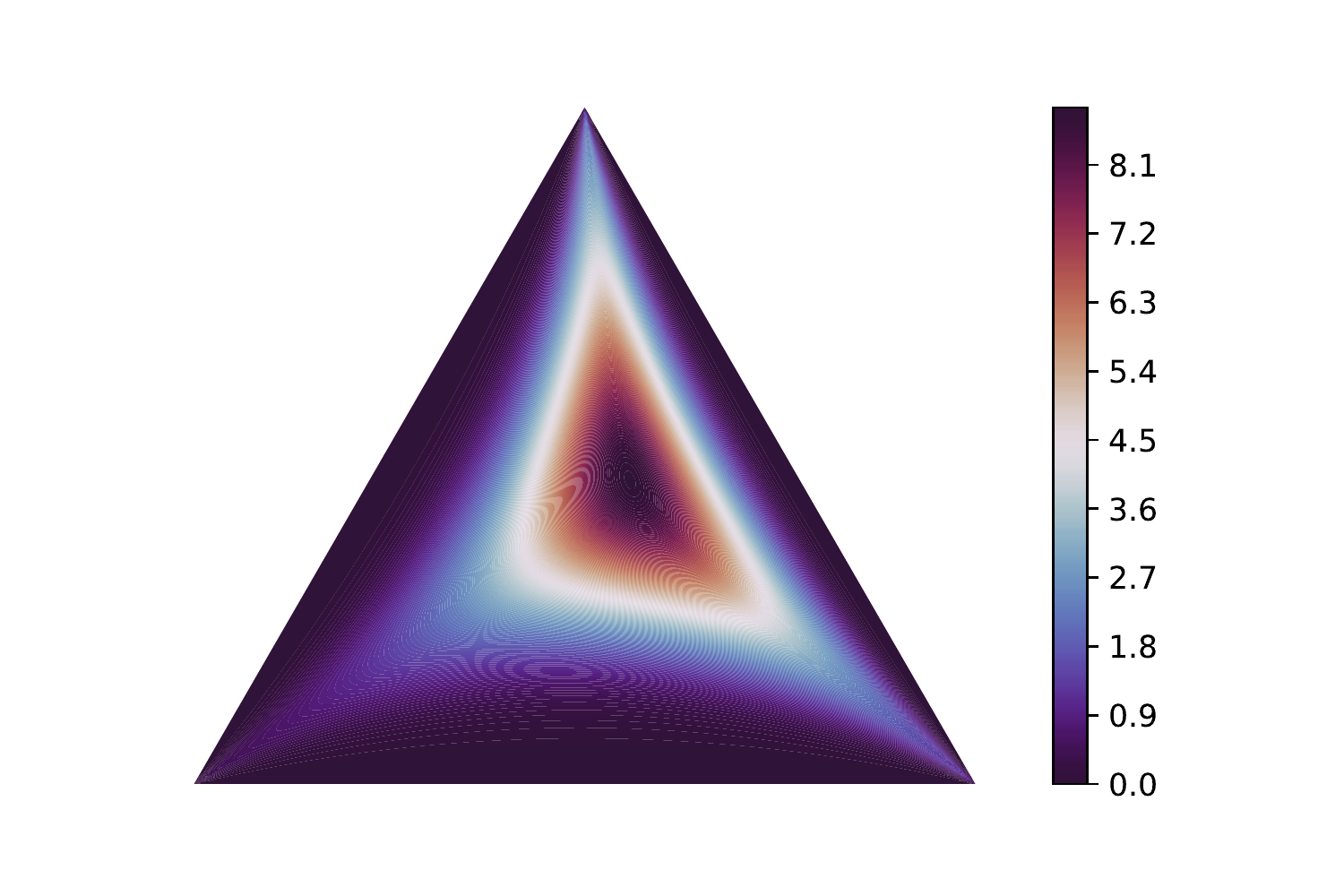}
        \caption{}
        \label{fig:sampling}
    \end{subfigure}\vspace{10pt}
    \hfill
    \caption{
    (a): Computational pipeline of the MarkerMap software package. Code to load and preprocess data, select markers, analyze and visualize results is provided. (b): Architecture of MarkerMap. Given input signals, a differentiable sampling process selects a global set of markers. In the supervised version of the method, the signal restricted to the selected markers is fed to a neural network that predicts labels. In the unsupervised version, the signal restricted to the selection is fed to a variational autoencoder that aims to reconstruct the original signal with no information of the label. The joint loss version uses a convex combination of the reconstruction loss and the classification loss. A circle represents a source of random inputs to be used for differentiable sampling (see the reparameterization trick \cite{kingma2013auto}). (c): Depiction of the Gumbel-Softmax distribution with probabilities $\left(\frac{1}{6},\frac{1}{3}, \frac{1}{2}\right)$ and temperature $\tau=2$ ( \emph{Methods}, Eq. 2). This distribution is used for sampling a set of most informative markers. The higher values indicate a higher probability of sampling a point there, so the center-right of this simplex has the highest probability of sampling a point there. As $\tau$ approaches 0, the density approaches a discrete density with higher intensity at the corners \cite{maddison_concrete}.}
    \label{fig:pipeline_architecture_sampling}
\end{figure*}

\section*{Results}

\subsection*{MarkerMap: learning relevant markers for scRNA-seq studies}

  We developed MarkerMap, a generative, deep learning marker selection framework which uses scRNA-seq data to extract a small number of genes which non-linearly combine to allow whole transcriptome reconstruction, without sacrificing accuracy on downstream prediction tasks. The input to MarkerMap is log normalized scRNA-seq data along with cell annotations such as cell type, spatial or intervention information, and a budget $k \in \mathbb{N}$. MarkerMap then outputs a set of $k$ genes (markers) which are most predictive of the output, together with a non-linear map for reconstructing the original gene expression space.

  Intuitively, MarkerMap computes feature importance scores for each gene in the input data using neural networks. These importance scores inform which genes are selected as representative of the input signal. MarkerMap then uses this reduced representation to compute an objective function predicting the given cell annotations (supervised; \emph{Methods}), reconstructing the full input signal (unsupervised; \emph{Methods}), or both (mixed strategy; \emph{Methods}). The selection step is probabilistic and is achieved through sampling from a discrete distribution which allows end-to-end optimization over the selection and predictive steps. The learnt mappings allow a) extracting the features most informative of a given clustering and b) generating full gene expression profiles when information from only the marker set is available.

  Technically, MarkerMap is an interpretable dimensionality reduction method based on the statistical framework of differentiable sampling optimization \cite{abid2019concrete-ae, maddison_concrete}. Targeted at addressing explainability tasks in machine learning, such methods have primarily been developed with text data in mind. Their performance has hence not been previously evaluated in a comprehensive way in the context of single cell studies. The relationship of MarkerMap with respect to these method and other previous approaches is discussed in \emph{Methods} and \emph{Tables 1, 2, and 3}.

  MarkerMap is available as well documented open-source software, along with tutorial and example workflows. 
The package provides a framework for custom designed feature selection methods along with metrics for evaluation (Figure 
\ref{fig:markermap_pipeline}).

\subsection*{Improving accuracy in supervised scRNA-seq studies}
We evaluated the performance of MarkerMap in the context of four publicly available scRNA-seq studies: Zeisel \cite{zeisel2015cell}, a CITE-seq technology based data set \cite{CITEseq}, a mouse brain scRNA-seq data set \cite{kleshchevnikov2020comprehensive},  and the Paul15 stem cell data set \cite{paul2015transcriptional} (see \emph{Methods} for a full description of the datasets and the data processing pipeline).

MarkerMap's performance is benchmarked against related non-linear approaches which, despite addressing related tasks, have not been  previously compared to one another. In detail, we considered the following feature selection baselines (\emph{Methods}):
LassoNet \cite{lemhadri2021lassonet}, SMaSH \cite{nelson2021smash}, and Concrete VAE \cite{abid2019concrete-ae}. We also adapted a continuous relaxation Gumbel-Softmax technique from \cite{xie2019reparameterizable} to allow for global feature selection, rather than local selection, in an effort to quantify the effect of the different sampling techniques on downstream clustering performance; we refer to this method as Global-Gate or Global-Gumbel VAE.

We report average misclassification and average F1 scores corresponding to a random forest classifier (Table \ref{tab:1}) and a k nearest neighbor classifier (Tab. \ref{tab:2}), across single cell data sets. We find that MarkerMap performs competitively with respect to these metrics, often improving on state of the art techniques. It is worth noting that, similar to empirical studies where dimensionality reduction is shown to improve the accuracy of downstream classification tasks \cite{nguyen2019ten}, the accuracy of the classifier trained only on features detected by MarkerMap  is often as good, or better, that that of the classifier trained on the full input.

\begin{table*}[t]
\begin{center}
\small
%\begin{tabular}{p{15mm}p{20mm}p{20mm}p{20mm}p{20mm}p{20mm}p{20mm}p{20mm}p{20mm}}
\begin{tabular}{cccccccc}
\textbf{Data sets}
&\textbf{Random Markers}
%& \multicolumn{1}{|p{2cm}|}{\centering GallusGallus \\ CC \\ t = 104}
&\textbf{SMaSH}
&\textbf{RankCorr}
%&  \multicolumn{1}{p{2cm}}{\centering\textbf{GlobalGate\\ VAE}}
& \multicolumn{1}{p{2cm}}{\centering \textbf{MarkerMap \\supervised}}
& \multicolumn{1}{p{2cm}}{\centering\textbf{Concrete\\ VAE}}
&\textbf{LassoNet} \\
\hline \\
\textbf{CITE-seq} & (0.813, 0.789) & (0.931, 0.919)& (0.869, 0.859)& \textbf{(0.939, 0.931)}& (0.821, 0.796) &	(0.938, 0.927)\\
\textbf{Mouse Brain}& (0.772, 0.748) & (0.974, 0.974)& (0.930, 0.929)& \textbf{(0.994, 0.994)}	 & (0.787, 0.765) &	(0.984, 0.984)\\
\textbf{Paul}& (0.544, 0.511) & (0.783, 0.771)& (0.665, 0.647)& (0.781, 0.769)	 & (0.542, 0.509) &(\textbf{0.787}, \textbf{0.777})\\
\textbf{Zeisel} &(0.724, 0.709) & (\textbf{0.958}, \textbf{0.953})& (0.944, 0.944)& (0.954, 0.953)	& (0.735, 0.722) &	(0.944, 0.942)
\end{tabular}
\caption{Average accuracy (first) and weighted F1 (second) scores across real single cell RNA-seq data sets, using a k-nearest neighbor classifier. All methods are instructed to select 50 markers. Higher values are better, and the top performer for each data set is bolded. Results are averaged over 50 runs.} \label{tab:2}
\end{center}
\end{table*}

Next, we evaluated how the average accuracy varies with the target number of selected markers (Fig. 2). We find that MarkerMap performs particularly well in a low selected marker regime, with less than 10\% marker selected. This can be particularly beneficial in applications like spatial transcriptomics where a small number of genes can be tagged for observation. For calibration, we also included a set of random markers (that we report as baseline). The random set of markers performed rather well, outperforming two of the methods considered -- Concrete VAE and Global-Gumbel VAE. We attribute the success of the random markers at classification to the high degree of correlation between features in biological studies. However, it is surprising that the sampling based baseline methods were outperformed by it. 
Next, we considered three variants of our method -- unsupervised, supervised, and joint (Tab. \ref{tab:1},\ref{tab:3}). Unsurprisingly, the supervised version performed best. The joint MarkerMap method was a close second, performing on par with the other top performers LassoNet and SMaSH. An attractive aspect distinguishing our method from SMaSH, in particular, is MarkerMap's additional reconstruction loss. This allows learning  markers that are both most predictive of cluster labels and best at reconstructing the full input data. This is particularly important in applications where feature collection is expensive or difficult. Finally, the unsupervised version of MarkerMap also had competitive performance. This version was trained without cluster information, hence suggesting that interpretable compression is possible for the biological study considered. When compared to approaches employing related sampling schemes
-- Concrete VAE and Global-Gumbel VAE (Tab. \ref{tab:3}), MarkerMap performs positively, suggesting that the differences in performance are largely due to parameter updating and aggregation across batches, rather than the sampling technique itself.

 \begin{figure*}[h]
\centering
    \includegraphics[width=0.6\textwidth]{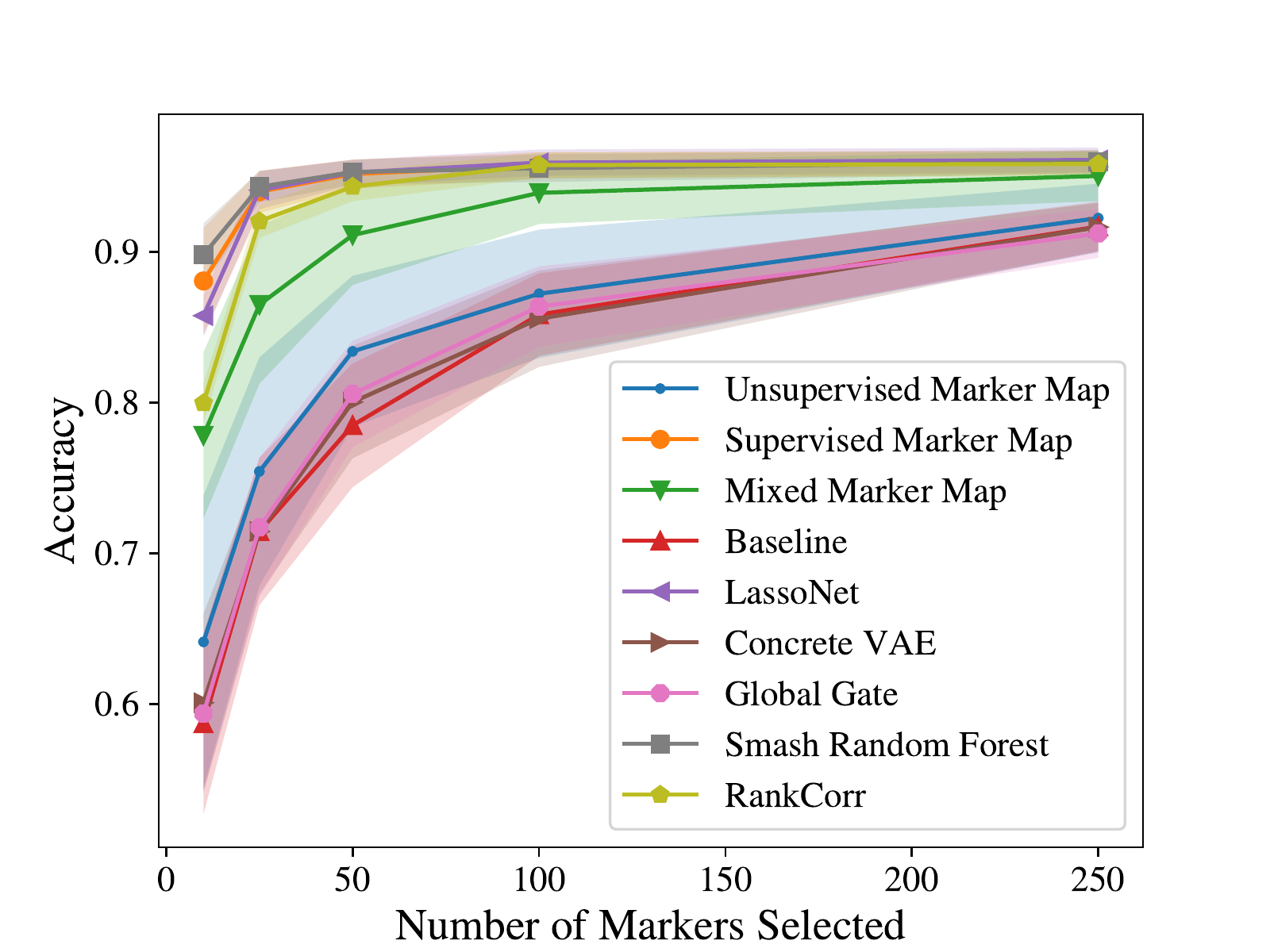}
    \caption{Mean accuracy and variance results from classifying cell types from a discrete set of markers chosen by the various methods on the Zeisel data set. The performance of the different methods is considered as a function of the number of markers selected with respect to a random forest classifier. Results are averaged over 50 runs. \label{fig:performance_plot_rf}}
\end{figure*}

\begin{figure*}[h]
    \centering
    \includegraphics[width=0.49\textwidth]{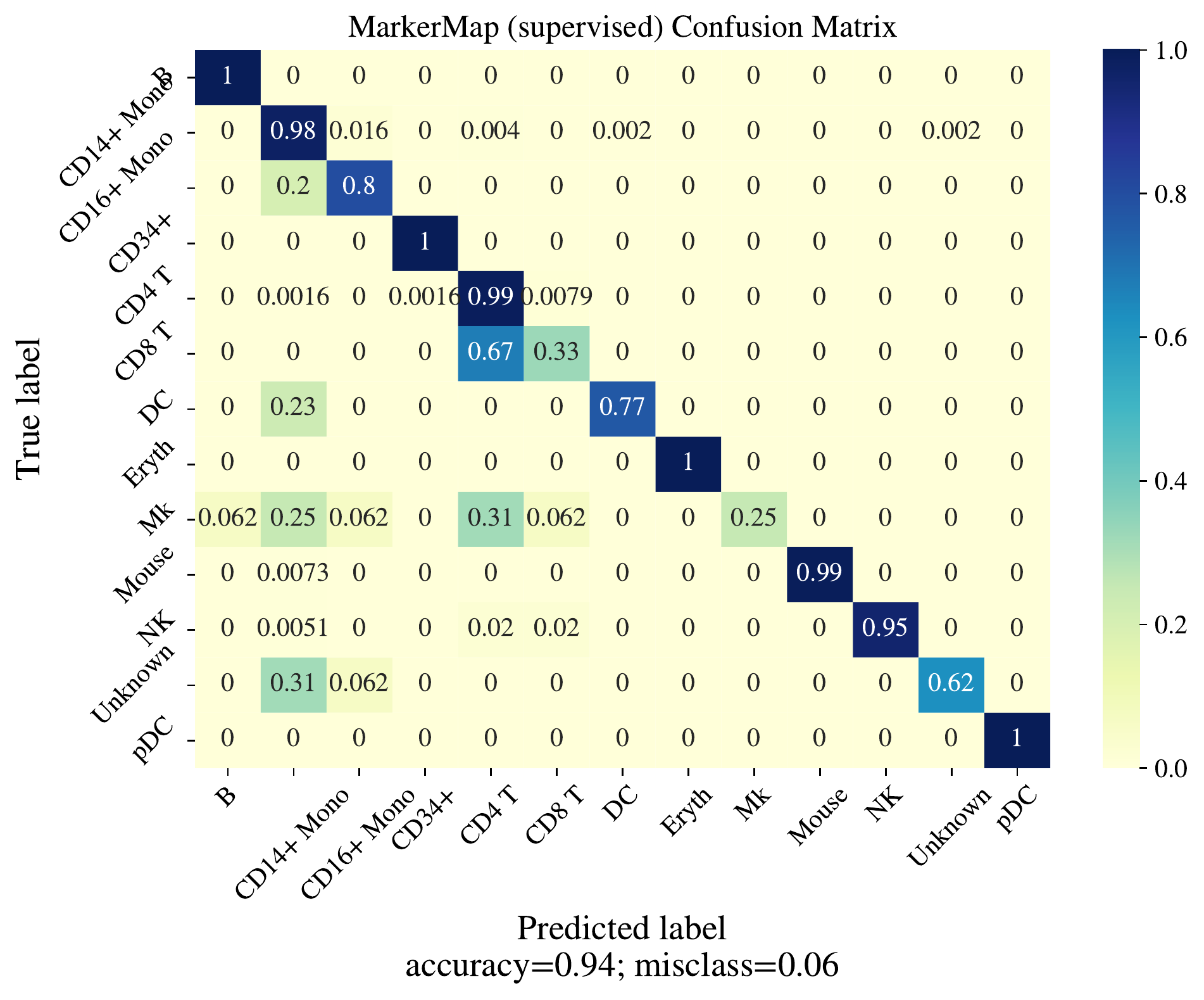}
    \includegraphics[width=0.49\textwidth]{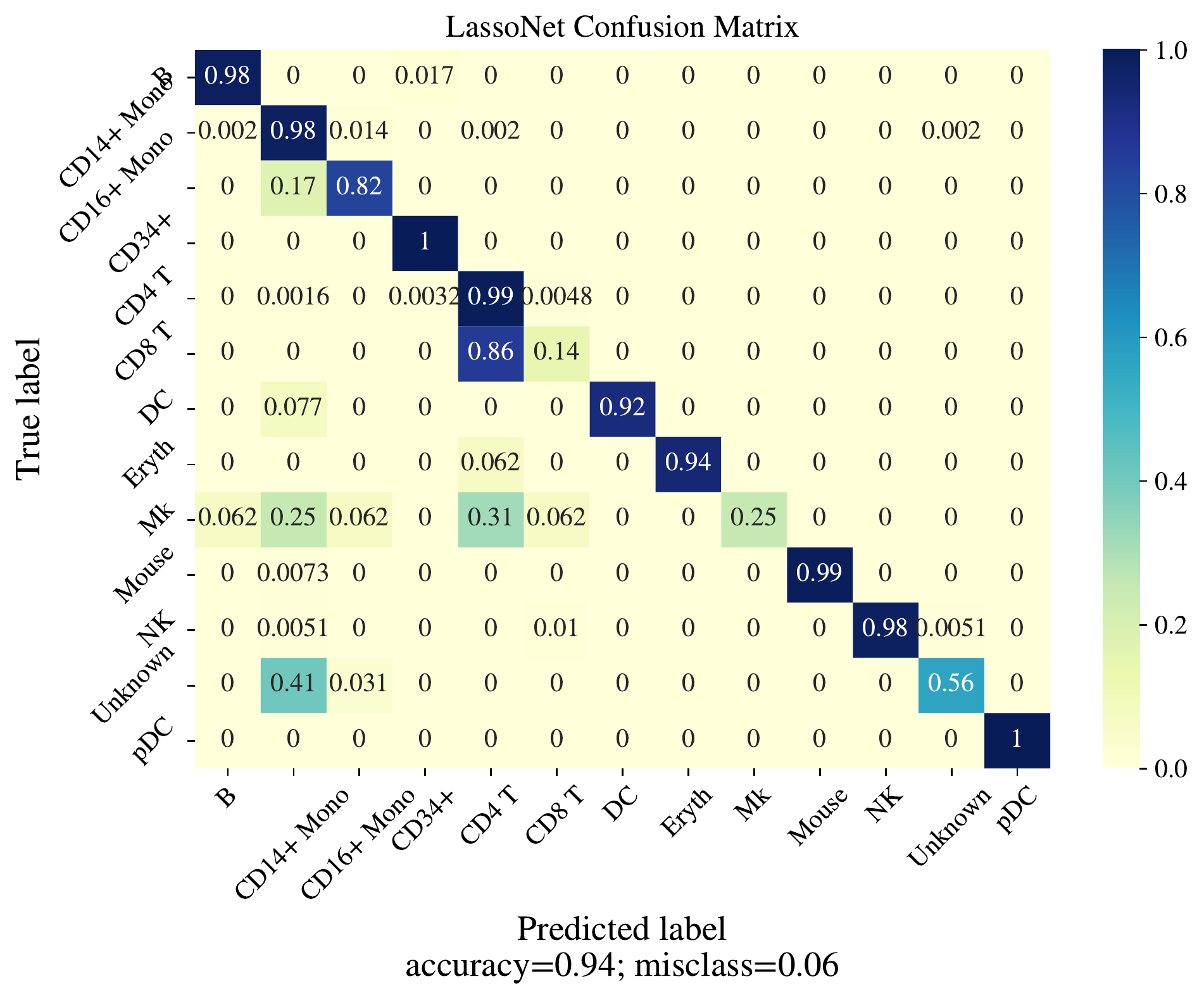}
    \includegraphics[width=0.49\textwidth]{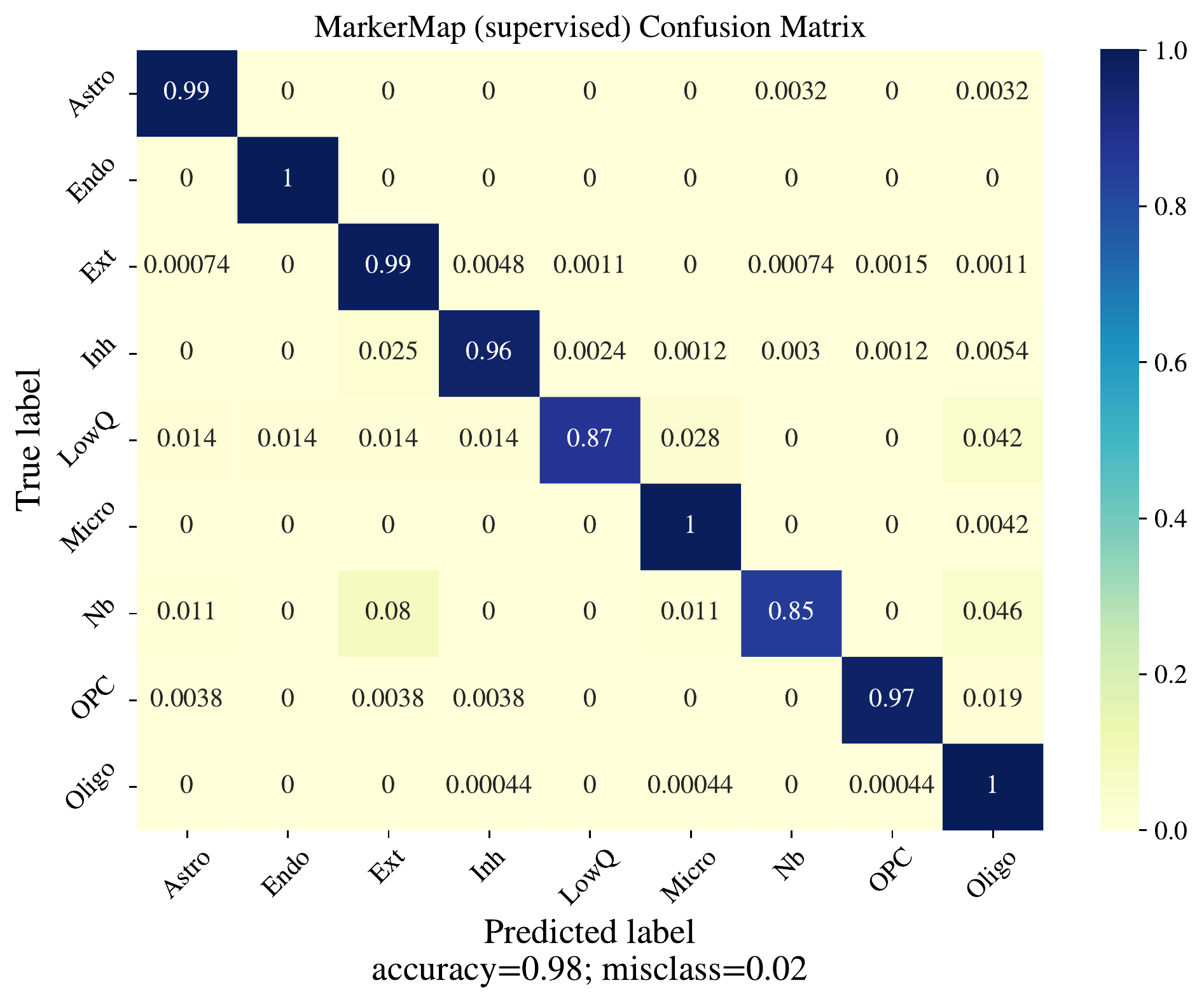}
    \includegraphics[width=0.49\textwidth]{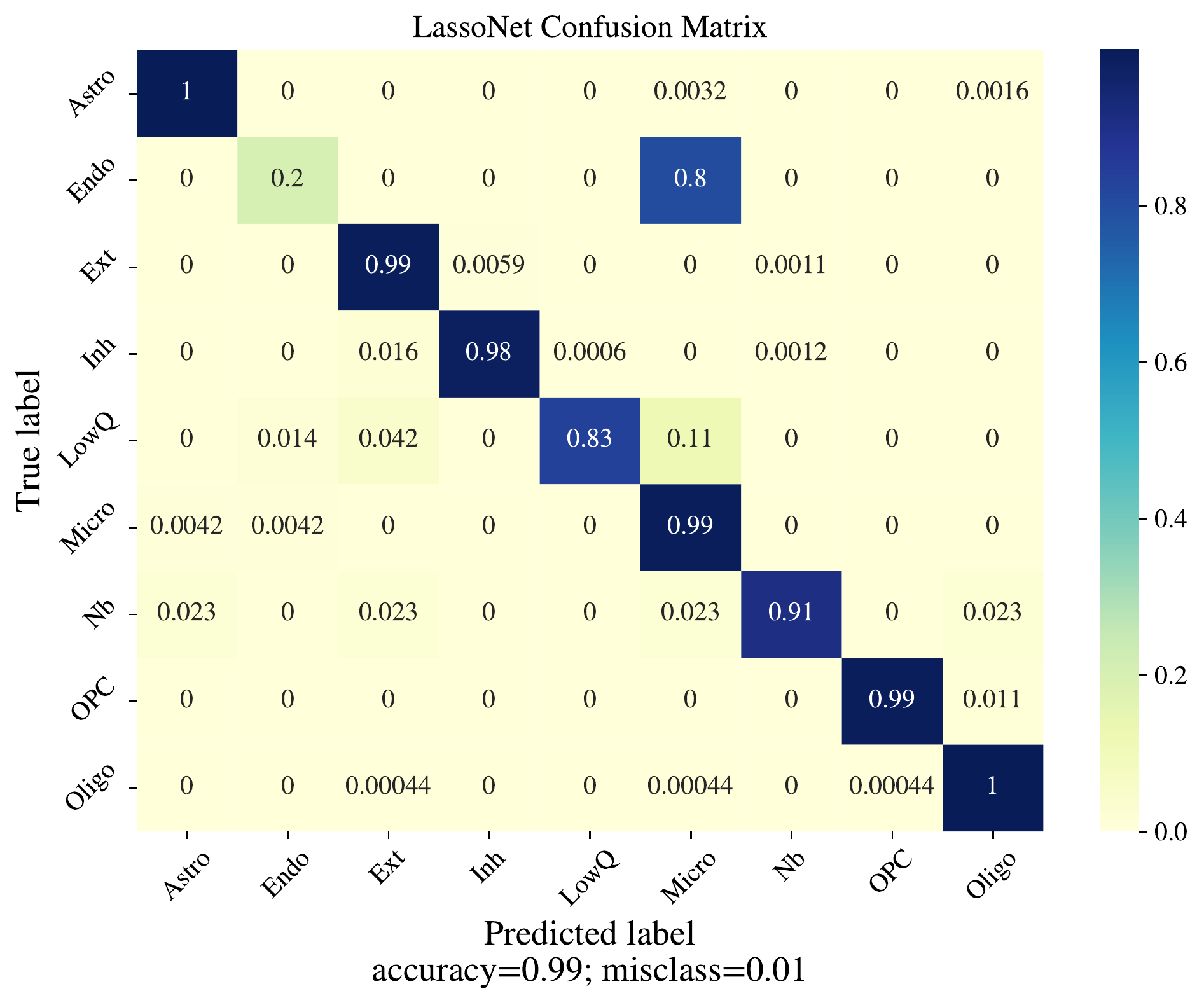}
%\vspace{1cm}
    \caption{Confusion matrices for the  top performing baselines on the CITE-seq data set: MarkerMap (left),  LassoNet (right). The bottom row is for the Mouse Brain data set. Each method was restricted to the selection of $50$ markers, and the classifier considered was a k-nearest neighbor classifier. }
    \label{fig:confusion_matrices}
\end{figure*}

Interestingly, even though MarkerMap and LassoNet present comparable overall misclassification errors, the individual cluster misclassification values are quite different (Fig. \ref{fig:confusion_matrices}). For example, in the CITE-seq data set, MarkerMap is slightly better at identifying the population of CD8 T and Eryth cells, while LassoNet is better at identifying the DC population and both methods have difficulties identifying Mk cells (Fig. \ref{fig:confusion_matrices}). Likewise, in the Mouse Brain data set, MarkerMap is better at identifying endothelial cells (End) and low quality cells (LowQ), while LassoNet is better at identifying neuroblastoma cells (Nb) (\ref{fig:confusion_matrices}). Given this, rather than advocating for a \emph{best} method for this task, we instead advocate for transparent, easy to use, top performing methods, which can pick up different signals from the data. 

%need to check if LowQ cells are low quality cells

\begin{table*}[h]
\begin{center}
\small
%\begin{tabular}{p{15mm}p{20mm}p{20mm}p{20mm}p{20mm}p{20mm}p{20mm}p{20mm}p{20mm}}
\begin{tabular}{cccccccc}
\textbf{Data sets}
& \multicolumn{1}{p{2cm}}{\centering \textbf{Global-Gumbel \\VAE}}

& \multicolumn{1}{p{2cm}}{\centering \textbf{MarkerMap \\unsupervised}}
%& \multicolumn{1}{|p{2cm}|}{\centering GallusGallus \\ CC \\ t = 104}
& \multicolumn{1}{p{2cm}}{\centering \textbf{MarkerMap \\supervised}}
& \multicolumn{1}{p{2cm}}{\centering \textbf{MarkerMap \\joint}}
& \multicolumn{1}{p{2cm}}{\centering \textbf{Concrete \\VAE}}
 \\

\hline \\
\textbf{CITE-seq} &
(0.873, 0.839)	& (0.888, 0.857) &	(\textbf{0.939, 0.922}) &	(0.928, 0.910)&	(0.873, 0.838)\\

\textbf{Mouse Brain} &
(0.854, 0.844)&	(0.985, 0.984)&	(\textbf{0.994, 0.994})&	(0.985, 0.984)&	(0.859, 0.850)
\\
\textbf{Paul} &
(0.615, 0.570)&	(0.857, 0.852)&	(\textbf{0.876, 0.873})&	(0.852, 0.847)&	(0.604, 0.556)
\\
\textbf{Zeisel} &
(0.806, 0.792) &	(0.834, 0.822)&	(\textbf{0.952, 0.951})&	(0.911, 0.906)&	(0.800, 0.785)
\end{tabular}
\caption{Comparison of sampling based methods. Average accuracy (first) and weighted F1 (second) scores across real single cell RNA-seq data sets, using a random Forest classifier. All methods are instructed to select 50 markers. Higher values are better, and the top performer for each data set is bolded. Results are averaged over 50 runs.} \label{tab:3}
\end{center}
\end{table*}

\subsection*{Learning representations which are robust to mislabeling}

Further, we investigated the effects of mislabelled training data on MarkerMap and different benchmarks. Cell type labels often come from different processing pipelines and can be error prone. Hence, marker selection methods ought to show robustness when the training labels are not completely accurate.  

\begin{figure*}
  \begin{subfigure}[a]{0.49\textwidth}
    \includegraphics[width=\textwidth]{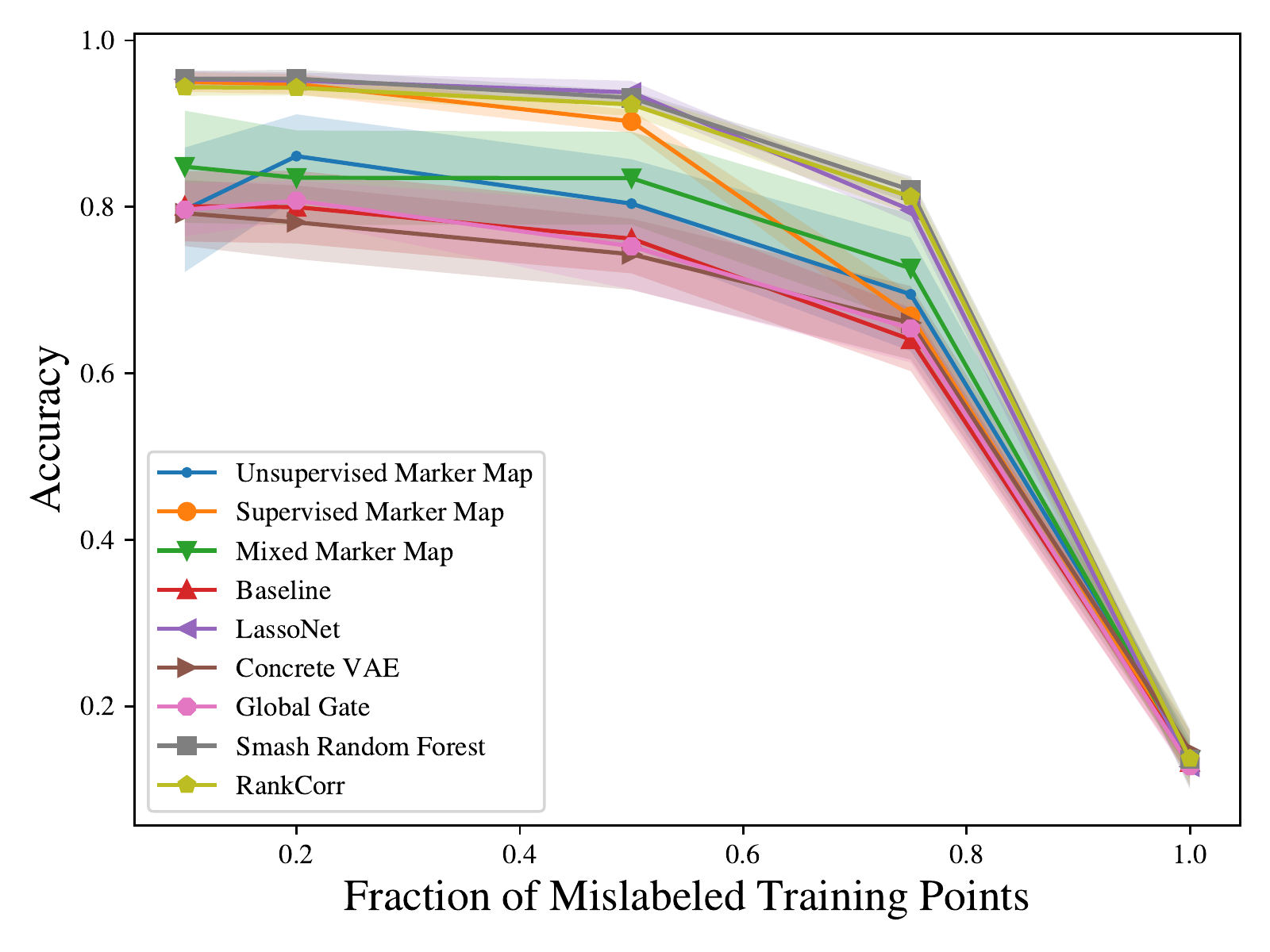}
    \caption{}
    \label{fig:noise_a}
  \end{subfigure}
  \begin{subfigure}[a]{0.49\textwidth}
    \includegraphics[width=\textwidth]{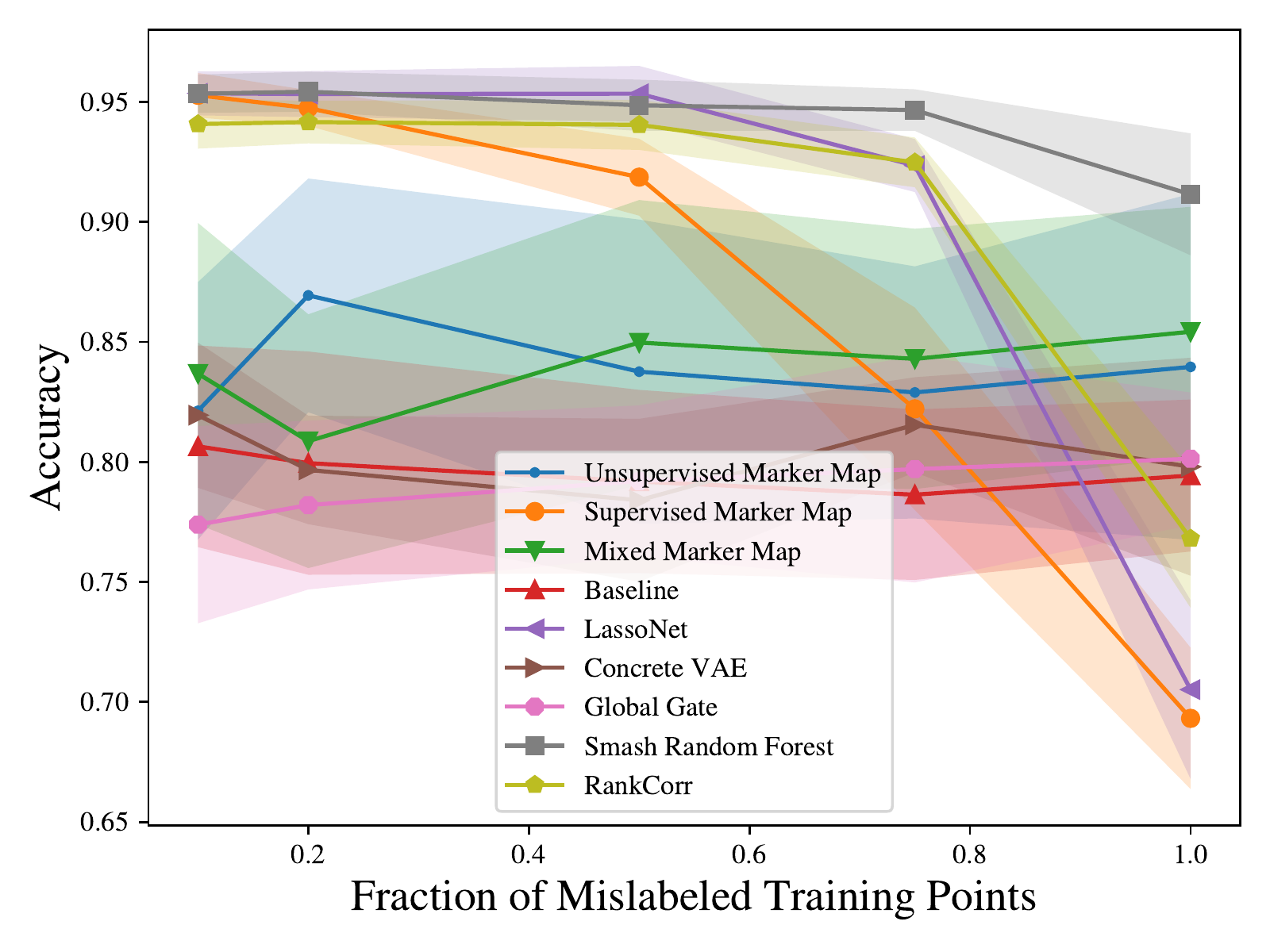}
    \caption{}
    \label{fig:noise_b}
  \end{subfigure}
  \caption{(a)  Robustness to label error in the training data with the Zeisel data set. (b) Robustness to label error when finding the markers, but not when training the simple classifier on those markers with the Zeisel data set. Each method selected 50 markers, and both plots are averaged over 10 runs.}
  \label{fig:noise}
\end{figure*}

To examine this effect we considered two experimental setups. First,  we replaced the labels of a fraction of the training set by a random label uniformly sampled over all the possible training labels (Fig. \eqref{fig:noise_a}). The misclassification rate was then measured only on the correctly labeled test data set. In this experiment, both the marker selection and the classifier were trained with incorrect labels so the performance decayed significantly when the fraction of misclassified points was large. Second, we also replaced the labels of a fraction of the training set by a random label uniformly sampled over all the possible training labels at the marker selection step, but the final classifier was trained on the correct labels on the (possibly incorrect) selected markers (Fig. \eqref{fig:noise_b}). This experiment suggests all top performing methods (MarkerMap, LassoNet) are similarly stable  with respect to noisy labels. The experiments also confirm that the performance of the unsupervised methods does not change, as they do not depend on input labels.

While the performance should be expected to deteriorate as the fraction of mislabelled training points increases, Figure \ref{fig:noise_a} shows that this happens \textit{slowly} for small label noise. While we do not have theoretical guarantees for such a robustness to mislabelling specifically for variational autoencoders, it is possible that this is an artifact of the general classification problem and the consistency of the estimator: \cite{LUGOSI199279} shows this to be the case of a nearest neighbor classifier under general conditions. Such a margin is large enough to accommodate realistic expectations of mislabelling error in data sets; we do however note that there may be more complex or systematic sources of error for which robustness may not hold. Figure \eqref{fig:noise_b}) echos the good performance of a set of random markers, when the number of markers is sufficiently large \cite{fischer2021many} and chosen to characterize a single cell type.

\begin{table*}[t]
\begin{center}
\small
%\begin{tabular}{p{15mm}p{20mm}p{20mm}p{20mm}p{20mm}p{20mm}p{20mm}p{20mm}p{20mm}}
\begin{tabular}{cccccccc}
\textbf{Data sets}
&\textbf{Random Markers}
%& \multicolumn{1}{|p{2cm}|}{\centering GallusGallus \\ CC \\ t = 104}
&\textbf{SMaSH}
&\textbf{RankCorr}
%&  \multicolumn{1}{p{2cm}}{\centering\textbf{GlobalGate\\ VAE}}
& \multicolumn{1}{p{2cm}}{\centering \textbf{MarkerMap \\supervised}}
& \multicolumn{1}{p{2cm}}{\centering\textbf{Concrete\\ VAE}}
&\textbf{LassoNet} \\
\hline \\
\textbf{CITE-seq} & (0.872, 0.837) & (0.935, 0.918)& (0.883, 0.856)& (\textbf{0.939}, \textbf{0.925})& (0.873, 0.838)& (0.937, 0.923)\\
\textbf{Mouse Brain}& (0.849, 0.840) & (0.982, 0.982)& (0.941, 0.940)& (\textbf{0.994}, \textbf{0.994}) & (0.859, 0.850)& (0.987, 0.987)\\
\textbf{Paul}& (0.607, 0.562) & (0.882, 0.879)& (0.774, 0.759)& (0.876, 0.873) & (0.604, 0.556)& (\textbf{0.885}, \textbf{881})\\
\textbf{Zeisel} &(0.785, 0.769) &(\textbf{0.953}, \textbf{0.952})& (0.943, 0.942)& (0.952, 0.951)& (0.800, 0.785)& (\textbf{0.953}, \textbf{0.952})
\end{tabular}
\caption{Average accuracy (first) and weighted F1 (second) scores across real single cell RNA-seq data sets, using a \textbf{Random Forest classifier}. All methods are instructed to select 50 markers. Higher values are better, and the top performer for each data set is bolded. Results are averaged over 50 runs.} \label{tab:1}
\end{center}
\end{table*}

\subsection*{Prospects for reconstruction in unsupervised settings}

\begin{figure*}[h]
    \centering
    \begin{subfigure}[b]{0.49\textwidth}
        \centering
        \includegraphics[width=\textwidth]{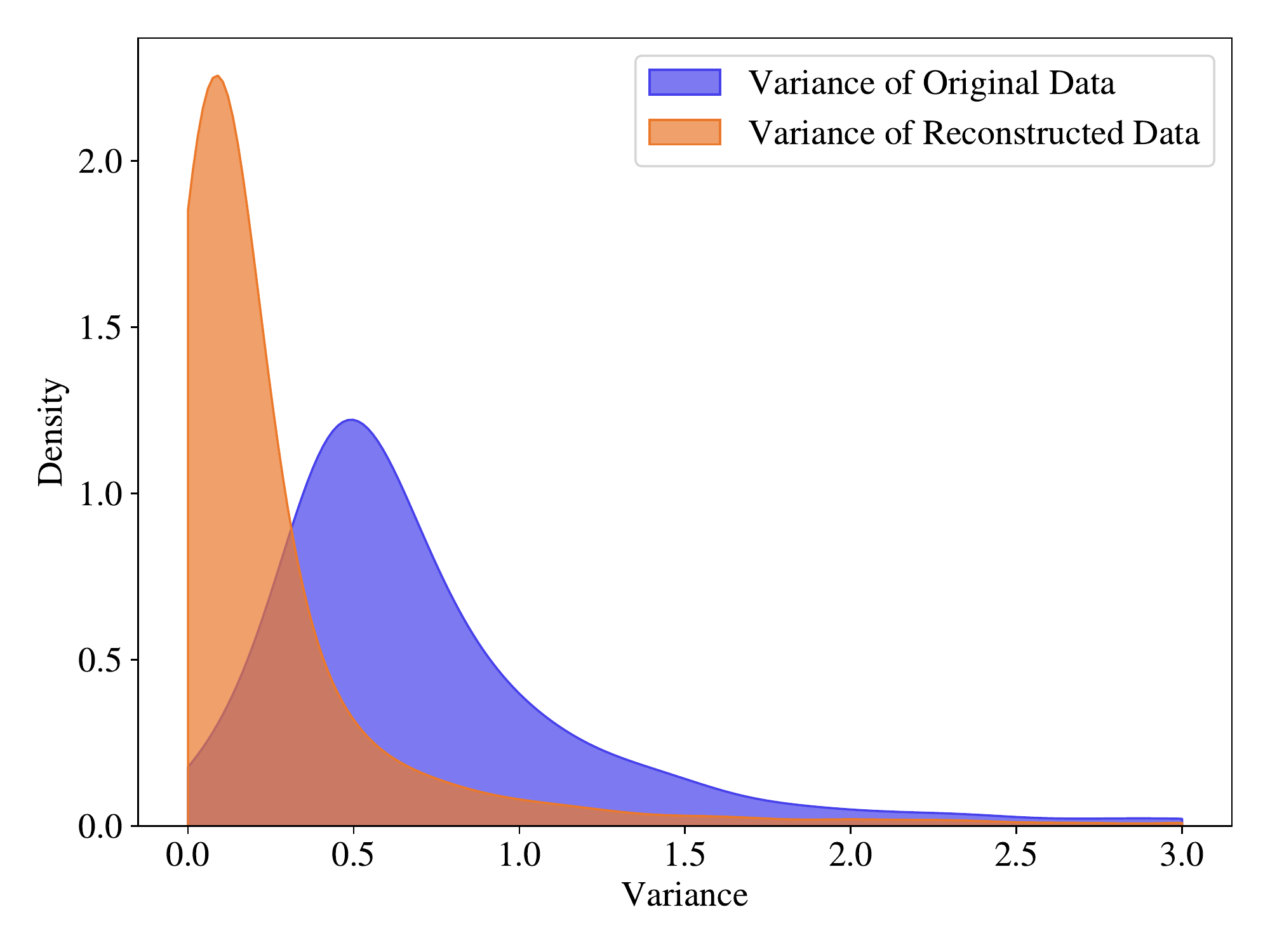}
        \caption{}
        \label{fig:recon_variances}
    \end{subfigure}
    \begin{subfigure}[b]{0.49\textwidth}
        \centering
        \includegraphics[width=\textwidth]{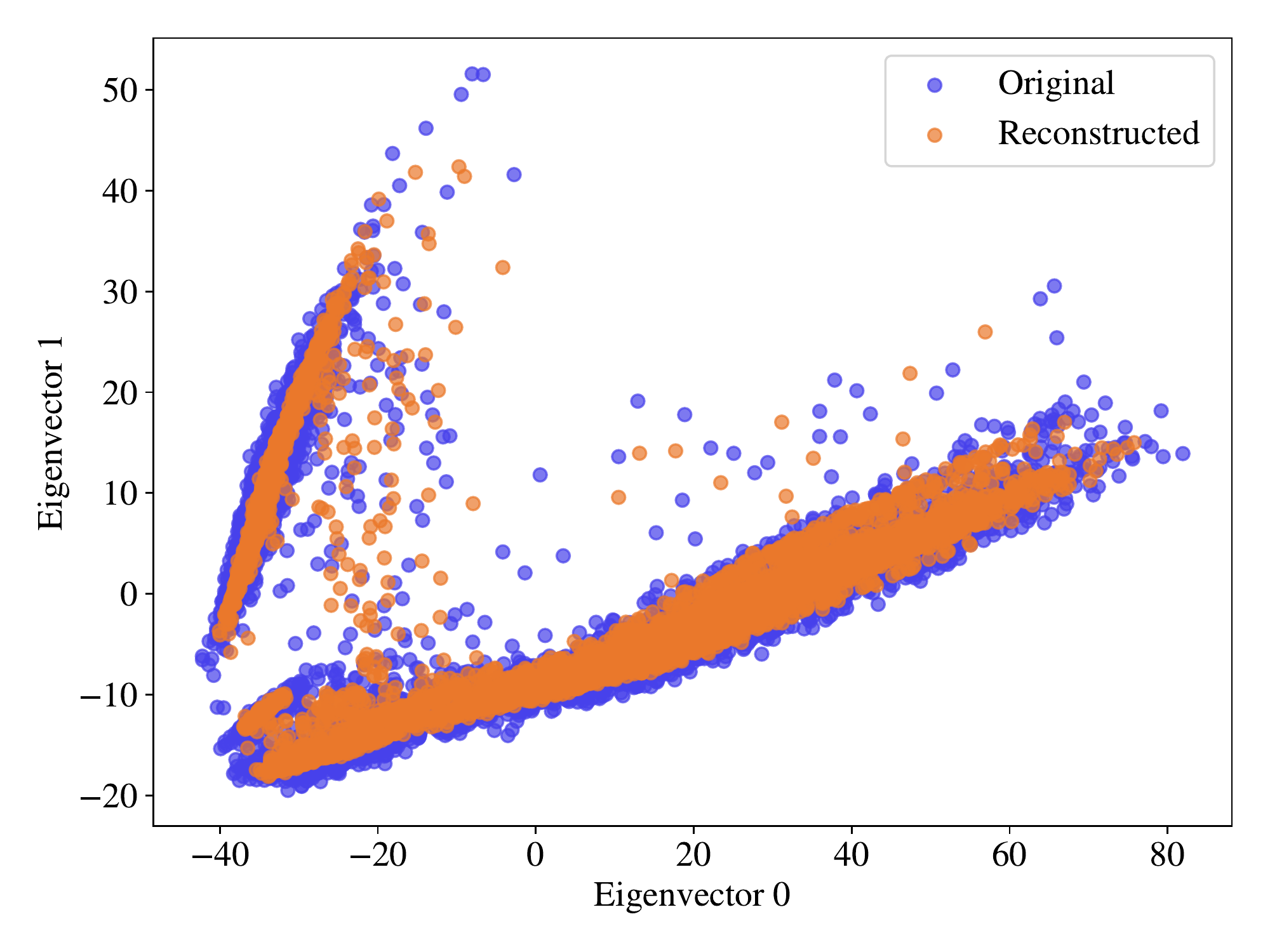}
        \caption{}
        \label{fig:pca_recon}
    \end{subfigure}
    \caption{Evaluating reconstructions. (a):  Density of gene expression variance values for data from the Mouse Brain data set: original samples and their corresponding MarkerMap reconstructions across all cell types. (b): PCA projections on the first two eigenvectors of the original data along with their reconstructed counterparts.}
    \label{fig:recon_evaluation}
\end{figure*}

As a generative model, MarkerMap allows the reconstruction of the full transcriptomic input from the selected set of most informative markers. To understand the limits of this recovery, we first quantified the  reconstruction quality by comparing distributional properties of the original and reconstructed datasets. Specifically, variances of genes from the reconstructed data were computed and compared to the variances of their counterparts in the original test data in a Mouse Brain data set, following unsupervised MarkerMap training with a 80\% - 20\% train-test split. The variances of the reconstructed data were lower than those of the original data (Fig. \ref{fig:recon_variances}). This is a common phenomenon for generative models obtained with variational autoencoders, known as variance shrinkage \cite{skafte2019reliable, akrami2020addressing}. To further visualize this, both test data and reconstructed data were projected onto the first two principle eigenvectors of the test data (Fig. \ref{fig:pca_recon}).

We further assessed whether, despite variance differences, the highly variable genes in the original data are recapitulated in the reconstructed one. To this end, two metrics for relative ranking were employed: the Jaccard Index and Spearman Rank Correlation Coefficient, $\rho$. Additionally, average $\ell_2$ distance between the reconstructed expression profiles and the original expression profiles were computed per cell type ( \emph{Evaluation Metrics} and \emph{Methods}). 

Each of these metrics were computed for both the reconstructed data from MarkerMap  and reconstructed data from a related generative model, scVI \cite{lopez2018}. The scVI model learns the parameters of a zero-inflated negative binomial distribution for modeling genes counts from scRNA-seq data \cite{lopez2018}. While both MarkerMap and scVI use a variational autoencoder framework for reconstruction, MarkerMap tries to reconstruct the full gene expression from the input of a small number of discrete markers, while scVI uses the full gene expression as input. In these experiments we used 50 markers for MarkerMap. Compared to scVI, MarkerMap generally scores worse on the variance metrics and better on the $
\ell_2$ distance  \ref{tab:mouse_brain_variances}. However, it should be noted that MarkerMap and scVI have slightly different goals that suggest that these results are appropriate. Unsupervised MarkerMap tries to find the best $k$ markers that optimally reconstruct the full data, while the scVI model learns a low dimensional manifold from which data is generated. A direction of future exploration is leveraging the differential sampling scheme of MarkerMap and the generative power of scVI to improve MarkerMap's reconstruction ability, while preserving its interpretability quality.

\begin{table*}[h]
    \begin{center}
        \small
        \begin{tabular}{c|ccc|ccc}
            & \textbf{MarkerMap} & &
            & \textbf{scVI} & \\
            \textbf{Cell Types} 
            &\textbf{Jaccard Index}
            &\textbf{Spearman $\rho$}
            &\textbf{$\ell_2$ Distance}
            &\textbf{Jaccard Index}
            &\textbf{Spearman $\rho$}
            &\textbf{$\ell_2$ Distance} \\
            \hline 
            \textbf{Astro} & 0.505 & 0.578 & 40.021 & 0.858 & 0.976 & 51.765 \\
            \textbf{Endo} & 0.162 & 0.110 & 42.404 & 0.240 & 0.265 & 50.352 \\
            \textbf{Ext} & 0.688 & 0.913 & 57.396 & 0.925 & 0.993 & 74.897 \\
            \textbf{Inh} & 0.663 & 0.869 & 56.244 & 0.905 & 0.988 & 73.628 \\
            \textbf{LowQ} & 0.551 & 0.722 & 52.773 & 0.690 & 0.859 & 66.035\\
            \textbf{Micro} & 0.340 & 0.351 & 35.379 & 0.762 & 0.945 & 44.239\\
            \textbf{Nb} & 0.351 & 0.438 & 42.272 & 0.634 & 0.848 & 53.200 \\
            \textbf{OPC} & 0.465 & 0.591 & 45.875 & 0.794 & 0.952 & 58.249 \\
            \textbf{Oligo} & 0.589 & 0.655 & 35.527 & 0.908 & 0.991 & 48.340 \\
            \hline 
            \textbf{All} & 0.738 & 0.908 & 48.234 & 0.939 & 0.994 & 63.354 \\
        \end{tabular}
        \caption{Comparing reconstruction quality metrics of the original Mouse Brain data against the reconstructed data broken down by cell type and overall. The left three columns are for MarkerMap reconstruction and the right three columns are for scVI reconstruction. Each value is the averaged over 5 random splits of the training and testing data. For the Jaccard Index and Spearman $\rho$, higher scores are better, while for $\ell_2$ distance, lower scores are better. MarkerMap map uses 50 markers, while scVI uses the full 4,581 genes.} 
        \label{tab:mouse_brain_variances}
    \end{center}
\end{table*}

\section*{Discussion}

 In this work we proposed MarkerMap, a data-driven, generative, neural network framework for interpretable feature selection. Given scRNA-seq data, we employed differentiable sampling methods to find a global set of genetic markers with competitive performance in downstream classification (of cell type) and reconstruction (of the entire transcriptome of an unseen test data). The supervised version selects the markers that maximize the label prediction accuracy. The unsupervised version selects markers that maximize the reconstruction accuracy of a variational autoencoder (with no label information). A mixed MarkerMap is also available, combining both label prediction and transcriptome reconstruction.
  
 We provide an extensive numerical benchmark of both supervised and unsupervised tools in the context of genetic marker selection on real single cell gene expression data sets. We show how the performance of all methods improve when the number of selected markers increases, and we analyze the stability of the methods in the presence of noisy labels. The baselines considered originated from different research communities, which have not been previously compared to one another despite addressing similar tasks. Moreover, we provide a pip installable python package, MarkerMap, that is easy to use, making it appropriate for experimental design.

Our experiments suggests that, even though differentiable sampling techniques based on properties of the Gumbel distribution are often suggested for interpretable machine learning tasks, they can underperform. Hence, the mathematically appealing, continuous relaxation procedure alone is not enough to explain why MarkerMap is competitive with respect to alternatives. Additional exploration, both experimental and theoretical, is required to understand this empirical result. In this work, we provide a competitive solution to feature selection in a real biological context. Most importantly, we provide a tool where related solutions from different fields can be compared to aid future research in this area. A promising future application of this tool is spatial trascriptomics.

\section*{Methods}

\subsection*{MarkerMap}

\subsubsection*{Introduction and Formulation}
MarkerMap is a generative method which belongs to the class of differentiable sampling techniques for subset selection \cite{maddison_concrete, abid2019concrete-ae, xie2019reparameterizable}. Existing differentiable sampling techniques aim to find local features that suit each input individually. These methods have been used for and are relevant to language contexts where the input is usually a sequence of variable length representing text. For example, in an online market setting, we might want to learn what specific words or group of words of a review are most predictive of the score associated with the review. Instead, MarkerMap seeks to find a global set of features (markers when referring to genes), amenable to the structure of scRNA-seq data, which results in optimization differences.

  In a nutshell, given high dimensional data points $\{x_i\}_{i=1}^n \subset \mathbb R^d$, arranged in a matrix $X\in \mathbb R^{n\times d}$, the feature selection problem aims to find a subset of coordinates $S\subset \{1,\ldots, d\}$, $|S|=K$, relevant to a given downstream task. For example, in sparse linear regression, data $X$ is used to predict responses $Y\in \mathbb R^n$ so that $Y \approx X \beta $ when only a small subset of the columns making up $X$ is relevant for the prediction. Similarly, in non-linear settings, the search is over a joint pair $(\beta, f)$, where is a non-linear function so that $Y \approx f(X \beta) $.

   Instead of optimizing for $\beta$, differentiable sampling methods assume informative samples are generated from a continuous distributions over a simplex with dimension equal to $K$, the number of features to be selected \cite{maddison_concrete, xie2019reparameterizable,abid2019concrete-ae, jang2016categorical}. This is accomplished through a selector layer. In detail, the selector layer contains $k=1,\ldots K$ nodes. Each node is associated with a $d$-dimensional real-valued vector $\gamma^{(k)}$ which governs the probability that a feature will be selected, whose entries $j$ are equal to:

\begin{align} \label{eq.logits}
\gamma_{j}^{(k)} = \frac{\exp((\log(\pi^{(k)}_j) + g_j^{(k)})/\tau)}{\sum_{s=1}^d \exp((\log(\pi^{(k)}_s) + g^{(k)}_s)/\tau)},
\end{align}

where $g_j^{(k)}$ are independent samples from a Gumbel distribution with location $0$ and scale $1$, $\tau$ is positive and real, and $\pi^{(k)}$ represent the class probabilities over a categorical distribution.  The $\gamma^{(k)}$ is a vector following a Gumbel-Softmax distribution, independently introduced by \cite{jang2016categorical} and \cite{maddison_concrete}. This distribution takes the form
\begin{equation}
  p_{\pi,\tau}(\gamma^{(1)},...\gamma^{(K)})=(K-1)!\tau^{K-1}\qty(\sum_{i=1}^K\frac{\pi^{(i)}}{\qty(\gamma^{(i)})^\tau})^{-K}\prod_{i=1}^K\qty(\frac{\pi^{(i)}}{\qty(\gamma^{(i)})^{\tau+1}}),
\end{equation}
and can be visualized over the $(K-1)$-dimensional simplex (Fig. \ref{fig:sampling}, for $K=3$).

We denote $\pi=(\pi^{(k)})_{k=1}^K$ and similarly $\gamma=(\gamma^{(k)})_{k=1}^K$. The number $\tau$ is referred to as temperature and the values $\log \pi $ are called logits. For an input $x_i$, each node $k$ of the selector layer outputs $x_i * \gamma^{(k)}$. As the temperature $\tau$ approaches $0$, $Pr(\gamma_{j}^{(k)} = 1) \to \pi_j^{(k)} / \sum_s \pi_s^{(k)}$, and only one feature of $x_i$ is selected \cite{abid2019concrete-ae}.

\subsubsection*{Optimization} 
Letting $p(x)$ be the probability distribution over the $d$-dimensional data $X$ and given a set of labels $Y$, MarkerMap learns: a) a subset of markers $S$ of size $K$, b) a reconstruction function $f_{\theta}: \mathbb{R}^K \to \mathbb{R}^d$, and c) a classifier $f_W: \mathbb{R}^K  \to \mathcal{Y}$.

   To learn these elements, the following empirical objective is optimized:
$$
\argmin_{S, \theta, W} \mathbb{E}_{p(x)}[\|f_{\theta} (x_S) - x \|_2 + \ell(f_W(x_S), y(x))]\\,
$$
where the first term optimizes signal reconstruction from a subset of markers $x_S$ and the second objective minimizes the expected classification risk, both over the unknown distribution $p(x)$ with respect to a loss function $\ell$. In practice, we consider the alternative empirical objective 
\begin{align} \label{eq.obj}
\underset{{S, \theta, W}}{\argmin}  \;\alpha \|f_{\theta} (X_S) - X \|_2  + (1- \alpha)\|(f_W(X_S), Y)\|_2,
\end{align}

  where $\alpha \in [0,1]$ serves to balance between a reconstruction loss and classification loss. MarkerMap considers three separate objectives: a supervised objective with $\alpha = 0$, an unsupervised objective with $\alpha = 1$, and a joint objective where $\alpha =0.5$. More generally, $\alpha$ can be treated as a tunable (but fixed) hyperparameter that weighs the reconstruction and classification terms in the optimization objective. Because full reconstruction is nominally a harder task it can be considered a bottleneck, since one can achieve low classification error without information about the entire gene expression. Thus, when $\alpha$ is small enough, the convergence of MarkerMap is dependent on the quality of the reconstruction. Depending on the user-specified goal, the three proposed values of $\alpha$ provide either a classifier ($\alpha=0$) which may be capable of selecting a smaller number of genes with good performance, a generative model ($\alpha=1$) which is capable of signal reconstruction possibly at the cost of additional markers needed, or both $(\alpha=0.5)$. One may choose a different value of $\alpha$ that is possibly data- or problem-specific.

  Optimizing this objective is difficult due to the combinatorial search over the subset $S$. We address this challenge heuristically by expanding on continuous sampling techniques \cite{xie2019reparameterizable} in a batch learning setting \cite{ioffe2015batch}. In a nutshell,  $b =1, 2, \ldots B$ batches are sampled without replacement from the data set $(X,Y)$. The selected features are then computed and aggregated across batches as follows:

\begin{enumerate}
    \item Instance-wise logits $\mathbf{\log\pi}_i^b = f_{\pi}(x_i)$ are generated  for each $x_i$ in the batch $b$, where $f_{\pi}$ is a neural network. Averaging them leads to an intermediate average batch logit $\mathbf{\log\pi}^b$.
    \item The average batch logits are computed by aggregating information from the current and previous batches, $\log\pi^{b} \leftarrow \beta \mathbf{\log\pi}^{b-1} + (1-\beta) \mathbf{\log\pi}^{b}$, $\beta \in (0, 1)$ much like the update for mean moment in BatchNorm \cite{ioffe2015batch}.
    \item The $K$ continuous \textit{d}-dimensional hot encoded vectors $\gamma^{(k),b} = (\gamma^{(k)}_{j})_{j=1,d}^b$ are generated from $\log\pi^b$ via continuous relaxation, see \eqref{eq.logits}.
    \item Each $\gamma^{(k),b}$ selects one of the $K$ features by element-wise multiplication $X_S^b = X^b \boxtimes \gamma^{b}$. 
    \item The resulting $X_S^b$ then becomes the input in a Variational-Autoencoder-like architecture, which includes a classifier loss as well as a reconstruction (Fig. \ref{fig:architecture} and \eqref{eq.obj}).
    \item All network weights are updated through stochastic gradient descent steps, following the optimization of the appropriate loss in \eqref{eq.obj} until convergence. The steps are repeated for $B$ timesteps, corresponding to the number of batches.
\end{enumerate}

\subsubsection*{Architecture}

The three main components of MarkerMap's architecture are the neural network $f_{\pi}$ for instance-wise logit generation, the task specific feed-forward network $f_{W}$ for classification, and the variational autoencoder $f_\theta$ for encoding and reconstruction. The neural network $f_{\pi}$ is an encoder with two hidden layers and a sampling layer performing relaxed subset sampling \cite{xie2019reparameterizable}. For supervised tasks, $f_{W}$ is represented by a decoder with one hidden layer. The encoder component of the variational autoencoder  $f_\theta$ has two hidden layers, while the Gaussian decoder has one hidden layer. All the hidden layers have the same size and are data set dependent, except for the Gaussian latent layer which has dimension 16 across experiments. The activation functions were chosen as follows: Leaky Rectified Linear Unit functions for hidden layers, identity transformation for the last layer of $f_\theta$ and softmax for the last layer of $f_W$. All activations were preceded by batch normalization in all hidden layers to mediate vanishing gradients.

\subsubsection*{Temperature annealing}

The temperature $\tau$ in \eqref{eq.logits} is a key parameter in the sampling procedure. It controls how fast the continuous encoding vectors $\gamma^{(k)}$ approach a true one-hot encoding. Low values of $\tau$ emulate true feature selection, while higher values of $\tau$ are more likely to extract linear combinations of features. However,  $0 <\tau < 1$ leads to inconsistent feature selection \cite{xie2019reparameterizable}. To mediate this issue, we used a temperature annealing scheme. First, we initialize $\tau_{\text{initial}} > 1$. This leads to gradients with less batch to batch variability and more diversity in feature selection, as $\boldsymbol{\gamma}^b$ will be more
diffuse. Second, we decay the temperature during training by a constant factor\cite{abid2019concrete-ae}. We found that setting $\tau_{\text{initial}} \geq 2$ with a decay factor leading to a $\tau_{\text{final}} \in (0.001, 0.1)$ resulted in good performance across all experiments.

\subsubsection*{Parameter initialization}

MarkerMap allows us to initialize the logits $\mathbf{\log\pi}^{b=0}$ with an informed guess of which markers are relevant. In the absence of prior information we initialize the logits as  $\mathbf{\log\pi}^{b=0} = \mathbf{1}c$, where $c$ is any constant. The weights of each linear layer are initialized using Kaiming initialization \cite{kaiming_initialization_2015}. The weights of the BatchNormalization layers are initialized as a vector of $\mathbf{1}$ for scaling and a vector of $\mathbf{0}$ for the biases.

  For backpropagation we use the Adam optimizer with a learning rate obtained via a learning rate finder \cite{smith2017cyclical}. A range of learning rates between 1e-8 and 0.001 are explored in linear intervals, with a minimum of 25 epochs and max of 100 epochs.
Training can end early when the average loss on the validation set does not decrease after 3 epochs.

  In all our experiments we randomly split the data in training (70\%), validation (10\%), and test sets (20\%). The batch size is 64 for all data sets. The quality of the markers did not seem to depend on batch size (with tested values of 32, 64, and 128 on Zeisel and Paul). We use a hidden layer size of 256 of Zeisel and Paul, 64 for CITEseq, and 500 for Mouse Brain.

\subsubsection*{Scalability}

Training MarkerMap on the 4,581 genes and 39,583 cells of the Mouse Brain data set (the largest considered) on public cloud GPUs resulted in a training time of 5 minutes for supervised classification tasks, and 15 minutes for unsupervised tasks. LassoNet performed similarly when the architecture (number of hidden layers and units) and batch sizes were chosen to be similar to those of MarkerMap. RankCorr and SMaSH achieved smaller training times, less than a minute, but require supervised signals.

\subsection*{Benchmarks}
We contrast MarkerMap against several subset selection methods. The methods have been introduced in different communities and have not been previously compared to one another.

\begin{itemize}
    \item LassoNet: A residual feed-forward network that makes use of an $\ell_1$ penalty on network weights in order to induce sparsity in selected features \cite{lemhadri2021lassonet}.
    \item Concrete VAE: a traditional VAE architecture that assumes a discrete distribution on latent parameters and performs inference using the formulation of the concrete distribution (also known as Gumbel-Softmax distribution) \cite{maddison_concrete}.
    \item Global-Gumbel VAE: adapted from \cite{xie2019reparameterizable}. A VAE architecture related to the Concrete VAE.
    \item Smash Random Forest: A classical Random Forest classification algorithm implemented in the \texttt{SMaSHpy} library\footnote{\url{https://pypi.org/project/smashpy/}}\cite{nelson2021smash}.
    \item RankCorr: A non-parametric marker selection method using (statistical) rank correlation, implemented in the \texttt{RankCorr} library\footnote{\url{https://github.com/ahsv/RankCorr}} \cite{vargo2020rank}.
\end{itemize}

\subsection*{Data sets}

We used publicly available real world data sets from established single cell analysis pipelines, where the problem of marker selection is of interest in the context of explaining cluster assignment.  In each data set, the labels correspond to cell types.

\textbf{Zeisel data set.}
The Zeisel data set contains data from $3,005$ cells and $4,000$ genes \cite{zeisel2015cell}. The cells were collected from the mouse somatosensory cortex (S1) and hippocampal CA1 region. The labels correspond to $7$ major cell types and where obtained though biclustering of the full gene expression data set.

\textbf{CITE-seq data set.}
Cellular Indexing of Transcriptomes and Epitopes by Sequencing (CITE-seq) is a single cell method that allows joint readouts from gene expression and proteins. The CITE-seq data set contains data from $8,617$ cells and $500$ genes \cite{CITEseq}. These cells correspond to major cord blood cells across $13$ cell types, obtained from the clustering of combined gene expression and protein read-out data, and not from the clustering of the original single cell data set alone. 

\textbf{Paul data set.}
The Paul data set \cite{paul2015transcriptional} consists of $2,730$ mouse bone marrow cells, collected with the MARS-seq protocol. Post processing, each cell contains $3,451$ genes. The Paul data set contains progenitor cells that are differentiating, hence the the data appear to follow a continuous trajectory. The associated outputs represent $10$ discrete cell types sampled along this trajectories. Hence, the cell types are are not well separated \cite{paul2015transcriptional}. After removing general genes and housekeeping genes, we are left with $3,113$ genes.

\textbf{Mouse Brain data set.}
This data set is a spatial transcriptomic data set, containing data from $40,532$ cells and $31,053$ genes from  diverse neuronal and glial cell types across stereotyped anatomical regions in the mouse brain \cite{kleshchevnikov2020comprehensive}. The output labels correspond to the major cell types identified by the authors. After some filtering of genes, we were working with $4,581$ genes. Training with the full data set
was not feasible for the unsupervised model on public cloud infrastructure.

\textbf{Data processing.}
The data were processed and filtered following \cite{CITEseq, nelson2021smash}. In particular, the data are sparse and normalized by a $log_2(1+X)$ transform. When evaluating the generative data, we forgo normalizing gene counts across cells and setting the mean to 0 and the variance to 1 of each gene. Instead, we only perform the $log_2(1+X)$ transform and then set the mean and variance of the entire data matrix $X$ to 0 and 1 respectively.

\subsection*{Evaluation Metrics}

Given $K$, most of the methods selected the top $K$ features informative of ground-truth labels. The exceptions, RankCorr and LassoNet, do not allow the selection of an exact number of features, as they rely on specifying a regularizer parameter that controls feature sparsity.  In those cases, we selected $K$ features by grid searching the regularizer that would get the desired number of features.

  For each baseline and data set, the selected features were then used as only input to a either a k-nearest neighbors classifier or a random forest classifier. For each data set, method and classifier type, we reported two quantities, the misclassification rate and a weighted F1 score, along with their corresponding confusion matrices. These quantities are defined as follows, for a number of ground truth clusters $c =1,2, \ldots C$.

\begin{itemize}
    \item \textbf{Average misclassification rate}. The misclassification rate of a given cluster is defined as 
    $$
    M_c = 1- \frac{TP_c}{TP_c + FP_c},
    $$
    where TP and FP correspond to the number of true positives and false positive predictions, respectively. We report the average misclassification $\frac{1}{C} \sum_c M_c$.
    \item \textbf{Average F1 score}. Per cluster, the F1 score is defined as
    $$
    F_c = \frac{2P_c R_c}{P_c + R_c},
    $$
    where $P_c$ and $R_c$ are the precision and recall of the classifier for a cluster $c$. We report the average F1 score $\frac{1}{C} \sum_c F_c$.
\end{itemize}

When evaluating the reconstructed data, we use the Jaccard Index, the Spearman Correlation Coefficient $\rho$, and the $\ell_2$ distance. Let $X \in \mathbb{R}^{n \times d}$ be our data as before, and let $\widetilde{X} \in \mathbb{R}^{n \times d}$ be the reconstructed data.
\begin{itemize}
    \item \textbf{Jaccard Index}. First we calculate the variances of each gene in the original data. Since each gene is a column of $X$, the variance of those columns is a $d$-length vector which we will denote $\sigma^2_{X}$. Next we find the rank vector of the variances, $R\qty(\sigma^2_X)$, where the largest variance is assigned 1, the second largest is assigned 2, and so on until the smallest variance is assigned $d$. We use the ranks to find the indices of the largest 20\% of the variances:
    \begin{equation*}
        I_X = \qty{i : R\qty(\sigma_X^2)[i] \leq \frac{d}{5}}
    \end{equation*}
    We follow the same process for the reconstructed data to get the set of indices $I_{\widetilde{X}}$. Finally, we calculate the Jaccard Index on these two sets of indices to determine their similarity \cite{jaccard1912}:
    \begin{equation*}
        J = \frac{\qty|I_{X} \cap I_{\widetilde{X}}|}{\qty|I_{X} \cup I_{\widetilde{X}}|}
    \end{equation*}
    The Jaccard Index ranges from 0 to 1, and higher values indicate that more of the highly variable genes from the original data are also highly variable in the reconstructed data.
    \item \textbf{Spearman Correlation Coefficient}. The Spearman correlation coefficient is exactly the Pearson correlation coefficient calculated on the ranks of a vector's values, rather than the raw values. Thus, we first calculate the rank vectors of the gene variances as we did for the Jaccard Index, $R\qty(\sigma^2_X)$ and $R\qty(\sigma^2_{\widetilde{X}})$. Finally we calculate the correlation coefficient:
    \begin{equation*}
        \rho = \frac{
            \text{cov}\qty(R\qty(\sigma^2_X),R\qty(\sigma^2_{\widetilde{X}}))
        }{
            \sigma_{R\qty(\sigma^2_X)} \sigma_{R\qty(\sigma^2_{\widetilde{X}})}
        }
    \end{equation*}
    where $\sigma_{R(\sigma^2_X)}$ and $\sigma_{R(\sigma^2_{\widetilde{X}})}$ are the standard deviations of the ranks of the original data and the reconstructed data respectively. This $\rho$ is the Spearman correlation coefficient --- values closer to one indicate higher similarity of the ranks of the gene variances.
    \item \textbf{$\ell_2$ Distance}. To calculate the $\ell_2$ distance, we take the average over all cells of the $\ell_2$ distance between the original cell and the reconstructed cell:
    \begin{equation*}
        \frac{1}{n}\sum_{i=1}^n \|x_i - \widetilde{x}_i\|_2
    \end{equation*}
    where $x_i$ is the $i^{th}$ row of $X$. Lower values indicate that the original data and reconstructed data are more similar.
\end{itemize}

\subsection*{Code availability}
The code is available as a Python package at https://github.com/Computational-Morphogenomics-Group/MarkerMap and on pip as ``markermap''. See \ref{fig:markermap_pipeline} for an overview of the package functionality. Code to easily load and pre-process the four data sets used in this paper are provided. Additional pre-processing can be done with the Scanpy package, and MarkerMap also provides functions to manage splitting the data into training and test sets. The package implements MarkerMap as well as Concrete VAE and Global Gate VAE. Additionally, it provides wrappers for LassoNet, SMaSH, and RankCorr to allow for easy benchmarking. All models select $k$ markers, which are then used for further tasks including visualizations.

\subsection*{Acknowledgments}
WG, GAK and SV were partially funded by ONR N00014-22-1-2126. SV is also partially funded by the
NSF–Simons Research Collaboration on the Mathematical and Scientific Foundations of Deep
Learning (MoDL) (NSF DMS 2031985), and the TRIPODS Institute for the Foundations of
Graph and Deep Learning at Johns Hopkins University. BD was supported by the Accelerate Programme for Scientific Discovery, funded by Schmidt Futures. The authors would like to thank Sinead Williamson for their comments on earlier versions of the manuscript.

\bibliographystyle{naturemag}
\bibliography{references}

\begin{thebibliography}{10}
\expandafter\ifx\csname url\endcsname\relax
  \def\url#1{\texttt{#1}}\fi
\expandafter\ifx\csname urlprefix\endcsname\relax\def\urlprefix{URL }\fi
\providecommand{\bibinfo}[2]{#2}
\providecommand{\eprint}[2][]{\url{#2}}

\bibitem{lohoff2020highly}
\bibinfo{author}{Lohoff, T.} \emph{et~al.}
\newblock \bibinfo{title}{Highly multiplexed spatially resolved gene expression
  profiling of mouse organogenesis}.
\newblock \emph{\bibinfo{journal}{bioRxiv}}  (\bibinfo{year}{2020}).

\bibitem{sladitschek2020morphoseq}
\bibinfo{author}{Sladitschek, H.~L.} \emph{et~al.}
\newblock \bibinfo{title}{Morphoseq: Full single-cell transcriptome dynamics up
  to gastrulation in a chordate}.
\newblock \emph{\bibinfo{journal}{Cell}}  (\bibinfo{year}{2020}).

\bibitem{codeluppi2018spatial}
\bibinfo{author}{Codeluppi, S.} \emph{et~al.}
\newblock \bibinfo{title}{Spatial organization of the somatosensory cortex
  revealed by cyclic sm{FISH}}.
\newblock \emph{\bibinfo{journal}{bioRxiv}} \bibinfo{pages}{276097}
  (\bibinfo{year}{2018}).

\bibitem{lubeck2014single}
\bibinfo{author}{Lubeck, E.}, \bibinfo{author}{Coskun, A.~F.},
  \bibinfo{author}{Zhiyentayev, T.}, \bibinfo{author}{Ahmad, M.} \&
  \bibinfo{author}{Cai, L.}
\newblock \bibinfo{title}{Single-cell in situ rna profiling by sequential
  hybridization}.
\newblock \emph{\bibinfo{journal}{Nature methods}}
  \textbf{\bibinfo{volume}{11}}, \bibinfo{pages}{360} (\bibinfo{year}{2014}).

\bibitem{chen2015spatially}
\bibinfo{author}{Chen, K.~H.}, \bibinfo{author}{Boettiger, A.~N.},
  \bibinfo{author}{Moffitt, J.~R.}, \bibinfo{author}{Wang, S.} \&
  \bibinfo{author}{Zhuang, X.}
\newblock \bibinfo{title}{Spatially resolved, highly multiplexed {RNA}
  profiling in single cells}.
\newblock \emph{\bibinfo{journal}{Science}} \textbf{\bibinfo{volume}{348}},
  \bibinfo{pages}{aaa6090} (\bibinfo{year}{2015}).

\bibitem{ke2013situ}
\bibinfo{author}{Ke, R.} \emph{et~al.}
\newblock \bibinfo{title}{In situ sequencing for rna analysis in preserved
  tissue and cells}.
\newblock \emph{\bibinfo{journal}{Nature methods}}
  \textbf{\bibinfo{volume}{10}}, \bibinfo{pages}{857--860}
  (\bibinfo{year}{2013}).

\bibitem{hotelling1933analysis}
\bibinfo{author}{Hotelling, H.}
\newblock \bibinfo{title}{Analysis of a complex of statistical variables into
  principal components.}
\newblock \emph{\bibinfo{journal}{Journal of educational psychology}}
  \textbf{\bibinfo{volume}{24}}, \bibinfo{pages}{417} (\bibinfo{year}{1933}).

\bibitem{kingma2013auto}
\bibinfo{author}{Kingma, D.~P.} \& \bibinfo{author}{Welling, M.}
\newblock \bibinfo{title}{Auto-encoding variational bayes}.
\newblock \emph{\bibinfo{journal}{arXiv preprint arXiv:1312.6114}}
  (\bibinfo{year}{2013}).

\bibitem{townes2019feature}
\bibinfo{author}{Townes, F.~W.}, \bibinfo{author}{Hicks, S.~C.},
  \bibinfo{author}{Aryee, M.~J.} \& \bibinfo{author}{Irizarry, R.~A.}
\newblock \bibinfo{title}{Feature selection and dimension reduction for
  single-cell rna-seq based on a multinomial model}.
\newblock \emph{\bibinfo{journal}{Genome biology}}
  \textbf{\bibinfo{volume}{20}}, \bibinfo{pages}{1--16} (\bibinfo{year}{2019}).

\bibitem{svensson2020interpretable}
\bibinfo{author}{Svensson, V.}, \bibinfo{author}{Gayoso, A.},
  \bibinfo{author}{Yosef, N.} \& \bibinfo{author}{Pachter, L.}
\newblock \bibinfo{title}{Interpretable factor models of single-cell rna-seq
  via variational autoencoders}.
\newblock \emph{\bibinfo{journal}{Bioinformatics}}
  \textbf{\bibinfo{volume}{36}}, \bibinfo{pages}{3418--3421}
  (\bibinfo{year}{2020}).

\bibitem{finak2015mast}
\bibinfo{author}{Finak, G.} \emph{et~al.}
\newblock \bibinfo{title}{Mast: a flexible statistical framework for assessing
  transcriptional changes and characterizing heterogeneity in single-cell rna
  sequencing data}.
\newblock \emph{\bibinfo{journal}{Genome biology}}
  \textbf{\bibinfo{volume}{16}}, \bibinfo{pages}{1--13} (\bibinfo{year}{2015}).

\bibitem{delaney2019combinatorial}
\bibinfo{author}{Delaney, C.} \emph{et~al.}
\newblock \bibinfo{title}{Combinatorial prediction of marker panels from
  single-cell transcriptomic data}.
\newblock \emph{\bibinfo{journal}{Molecular systems biology}}
  \textbf{\bibinfo{volume}{15}}, \bibinfo{pages}{e9005} (\bibinfo{year}{2019}).

\bibitem{ibrahim2019genesorter}
\bibinfo{author}{Ibrahim, M.~M.} \& \bibinfo{author}{Kramann, R.}
\newblock \bibinfo{title}{Genesorter: feature ranking in clustered single cell
  data}.
\newblock \emph{\bibinfo{journal}{BioRxiv}} \bibinfo{pages}{676379}
  (\bibinfo{year}{2019}).

\bibitem{dumitrascu2021optimal}
\bibinfo{author}{Dumitrascu, B.}, \bibinfo{author}{Villar, S.},
  \bibinfo{author}{Mixon, D.~G.} \& \bibinfo{author}{Engelhardt, B.~E.}
\newblock \bibinfo{title}{Optimal marker gene selection for cell type
  discrimination in single cell analyses}.
\newblock \emph{\bibinfo{journal}{Nature communications}}
  \textbf{\bibinfo{volume}{12}}, \bibinfo{pages}{1--8} (\bibinfo{year}{2021}).

\bibitem{vargo2020rank}
\bibinfo{author}{Vargo, A.~H.} \& \bibinfo{author}{Gilbert, A.~C.}
\newblock \bibinfo{title}{A rank-based marker selection method for high
  throughput scrna-seq data}.
\newblock \emph{\bibinfo{journal}{BMC bioinformatics}}
  \textbf{\bibinfo{volume}{21}}, \bibinfo{pages}{1--51} (\bibinfo{year}{2020}).

\bibitem{nelson2021smash}
\bibinfo{author}{Nelson, M.~E.}, \bibinfo{author}{Riva, S.~G.} \&
  \bibinfo{author}{Cvejic, A.}
\newblock \bibinfo{title}{Smash: A scalable, general marker gene identification
  framework for single-cell rna sequencing and spatial transcriptomics}.
\newblock \emph{\bibinfo{journal}{bioRxiv}}  (\bibinfo{year}{2021}).

\bibitem{conrad2017sparse}
\bibinfo{author}{Conrad, T.~O.} \emph{et~al.}
\newblock \bibinfo{title}{Sparse proteomics analysis--a compressed
  sensing-based approach for feature selection and classification of
  high-dimensional proteomics mass spectrometry data}.
\newblock \emph{\bibinfo{journal}{BMC bioinformatics}}
  \textbf{\bibinfo{volume}{18}}, \bibinfo{pages}{1--20} (\bibinfo{year}{2017}).

\bibitem{shrikumar2017learning}
\bibinfo{author}{Shrikumar, A.}, \bibinfo{author}{Greenside, P.} \&
  \bibinfo{author}{Kundaje, A.}
\newblock \bibinfo{title}{Learning important features through propagating
  activation differences}.
\newblock In \emph{\bibinfo{booktitle}{International Conference on Machine
  Learning}}, \bibinfo{pages}{3145--3153} (\bibinfo{organization}{PMLR},
  \bibinfo{year}{2017}).

\bibitem{mcwhirter2019squeezefit}
\bibinfo{author}{McWhirter, C.}, \bibinfo{author}{Mixon, D.~G.} \&
  \bibinfo{author}{Villar, S.}
\newblock \bibinfo{title}{Squeezefit: Label-aware dimensionality reduction by
  semidefinite programming}.
\newblock \emph{\bibinfo{journal}{IEEE Transactions on Information Theory}}
  \textbf{\bibinfo{volume}{66}}, \bibinfo{pages}{3878--3892}
  (\bibinfo{year}{2019}).

\bibitem{liang2021single}
\bibinfo{author}{Liang, S.} \emph{et~al.}
\newblock \bibinfo{title}{Single-cell manifold-preserving feature selection for
  detecting rare cell populations}.
\newblock \emph{\bibinfo{journal}{Nature Computational Science}}
  \textbf{\bibinfo{volume}{1}}, \bibinfo{pages}{374--384}
  (\bibinfo{year}{2021}).

\bibitem{yang2021feature}
\bibinfo{author}{Yang, P.}, \bibinfo{author}{Huang, H.} \&
  \bibinfo{author}{Liu, C.}
\newblock \bibinfo{title}{Feature selection revisited in the single-cell era}.
\newblock \emph{\bibinfo{journal}{Genome biology}}
  \textbf{\bibinfo{volume}{22}}, \bibinfo{pages}{1--17} (\bibinfo{year}{2021}).

\bibitem{pullin2022comparison}
\bibinfo{author}{Pullin, J.~M.} \& \bibinfo{author}{McCarthy, D.~J.}
\newblock \bibinfo{title}{A comparison of marker gene selection methods for
  single-cell rna sequencing data}.
\newblock \emph{\bibinfo{journal}{bioRxiv}}  (\bibinfo{year}{2022}).

\bibitem{tibshirani1996regression}
\bibinfo{author}{Tibshirani, R.}
\newblock \bibinfo{title}{Regression shrinkage and selection via the lasso}.
\newblock \emph{\bibinfo{journal}{Journal of the Royal Statistical Society:
  Series B (Methodological)}} \textbf{\bibinfo{volume}{58}},
  \bibinfo{pages}{267--288} (\bibinfo{year}{1996}).

\bibitem{mahoney2009cur}
\bibinfo{author}{Mahoney, M.~W.} \& \bibinfo{author}{Drineas, P.}
\newblock \bibinfo{title}{Cur matrix decompositions for improved data
  analysis}.
\newblock \emph{\bibinfo{journal}{Proceedings of the National Academy of
  Sciences}} \textbf{\bibinfo{volume}{106}}, \bibinfo{pages}{697--702}
  (\bibinfo{year}{2009}).

\bibitem{lemhadri2021lassonet}
\bibinfo{author}{Lemhadri, I.}, \bibinfo{author}{Ruan, F.},
  \bibinfo{author}{Abraham, L.} \& \bibinfo{author}{Tibshirani, R.}
\newblock \bibinfo{title}{Lassonet: A neural network with feature sparsity}.
\newblock \emph{\bibinfo{journal}{Journal of Machine Learning Research}}
  \textbf{\bibinfo{volume}{22}}, \bibinfo{pages}{1--29} (\bibinfo{year}{2021}).

\bibitem{maddison_concrete}
\bibinfo{author}{Maddison, C.~J.}, \bibinfo{author}{Mnih, A.} \&
  \bibinfo{author}{Teh, Y.~W.}
\newblock \bibinfo{title}{The concrete distribution: {A} continuous relaxation
  of discrete random variables}.
\newblock In \emph{\bibinfo{booktitle}{5th International Conference on Learning
  Representations, {ICLR} 2017, Toulon, France, April 24-26, 2017, Conference
  Track Proceedings}} (\bibinfo{year}{2017}).

\bibitem{xie2019reparameterizable}
\bibinfo{author}{Xie, S.~M.} \& \bibinfo{author}{Ermon, S.}
\newblock \bibinfo{title}{Reparameterizable subset sampling via continuous
  relaxations}.
\newblock In \emph{\bibinfo{booktitle}{International Joint Conference on
  Artificial Intelligence}} (\bibinfo{year}{2019}).

\bibitem{abid2019concrete-ae}
\bibinfo{author}{Abid, A.}, \bibinfo{author}{Balin, M.~F.} \&
  \bibinfo{author}{Zou, J.}
\newblock \bibinfo{title}{Concrete autoencoders for differentiable feature
  selection and reconstruction}.
\newblock \emph{\bibinfo{journal}{arXiv preprint arXiv:1901.09346}}
  (\bibinfo{year}{2019}).

\bibitem{jang2016categorical}
\bibinfo{author}{Jang, E.}, \bibinfo{author}{Gu, S.} \& \bibinfo{author}{Poole,
  B.}
\newblock \bibinfo{title}{Categorical reparameterization with gumbel-softmax}.
\newblock \emph{\bibinfo{journal}{arXiv preprint arXiv:1611.01144}}
  (\bibinfo{year}{2016}).

\bibitem{zeisel2015cell}
\bibinfo{author}{Zeisel, A.} \emph{et~al.}
\newblock \bibinfo{title}{Cell types in the mouse cortex and hippocampus
  revealed by single-cell {RNA}-seq}.
\newblock \emph{\bibinfo{journal}{Science}} \textbf{\bibinfo{volume}{347}},
  \bibinfo{pages}{1138--1142} (\bibinfo{year}{2015}).

\bibitem{CITEseq}
\bibinfo{author}{Stoeckius, M.} \emph{et~al.}
\newblock \bibinfo{title}{Simultaneous epitope and transcriptome measurement in
  single cells}.
\newblock \emph{\bibinfo{journal}{Nature methods}}
  \textbf{\bibinfo{volume}{14}}, \bibinfo{pages}{865} (\bibinfo{year}{2017}).

\bibitem{kleshchevnikov2020comprehensive}
\bibinfo{author}{Kleshchevnikov, V.} \emph{et~al.}
\newblock \bibinfo{title}{Comprehensive mapping of tissue cell architecture via
  integrated single cell and spatial transcriptomics}.
\newblock \emph{\bibinfo{journal}{bioRxiv}}  (\bibinfo{year}{2020}).

\bibitem{paul2015transcriptional}
\bibinfo{author}{Paul, F.} \emph{et~al.}
\newblock \bibinfo{title}{Transcriptional heterogeneity and lineage commitment
  in myeloid progenitors}.
\newblock \emph{\bibinfo{journal}{Cell}} \textbf{\bibinfo{volume}{163}},
  \bibinfo{pages}{1663--1677} (\bibinfo{year}{2015}).

\bibitem{nguyen2019ten}
\bibinfo{author}{Nguyen, L.~H.} \& \bibinfo{author}{Holmes, S.}
\newblock \bibinfo{title}{Ten quick tips for effective dimensionality
  reduction}.
\newblock \emph{\bibinfo{journal}{PLoS computational biology}}
  \textbf{\bibinfo{volume}{15}}, \bibinfo{pages}{e1006907}
  (\bibinfo{year}{2019}).

\bibitem{LUGOSI199279}
\bibinfo{author}{Lugosi, G.}
\newblock \bibinfo{title}{Learning with an unreliable teacher}.
\newblock \emph{\bibinfo{journal}{Pattern Recognition}}
  \textbf{\bibinfo{volume}{25}}, \bibinfo{pages}{79--87}
  (\bibinfo{year}{1992}).
\newblock
  \urlprefix\url{https://www.sciencedirect.com/science/article/pii/0031320392900087}.

\bibitem{fischer2021many}
\bibinfo{author}{Fischer, S.} \& \bibinfo{author}{Gillis, J.}
\newblock \bibinfo{title}{How many markers are needed to robustly determine a
  cell's type?}
\newblock \emph{\bibinfo{journal}{Iscience}} \textbf{\bibinfo{volume}{24}},
  \bibinfo{pages}{103292} (\bibinfo{year}{2021}).

\bibitem{skafte2019reliable}
\bibinfo{author}{Skafte, N.}, \bibinfo{author}{J{\o}rgensen, M.} \&
  \bibinfo{author}{Hauberg, S.}
\newblock \bibinfo{title}{Reliable training and estimation of variance
  networks}.
\newblock \emph{\bibinfo{journal}{Advances in Neural Information Processing
  Systems}} \textbf{\bibinfo{volume}{32}} (\bibinfo{year}{2019}).

\bibitem{akrami2020addressing}
\bibinfo{author}{Akrami, H.}, \bibinfo{author}{Joshi, A.~A.},
  \bibinfo{author}{Aydore, S.} \& \bibinfo{author}{Leahy, R.~M.}
\newblock \bibinfo{title}{Addressing variance shrinkage in variational
  autoencoders using quantile regression}.
\newblock \emph{\bibinfo{journal}{arXiv preprint arXiv:2010.09042}}
  (\bibinfo{year}{2020}).

\bibitem{lopez2018}
\bibinfo{author}{Lopez, R.}, \bibinfo{author}{Regier, J.},
  \bibinfo{author}{Cole, M.~B.}, \bibinfo{author}{Jordan, M.~I.} \&
  \bibinfo{author}{Yosef, N.}
\newblock \bibinfo{title}{{Deep generative modeling for single-cell
  transcriptomics}}.
\newblock \emph{\bibinfo{journal}{Nature methods}}
  \textbf{\bibinfo{volume}{15}}, \bibinfo{pages}{1053--1058}
  (\bibinfo{year}{2018}).

\bibitem{ioffe2015batch}
\bibinfo{author}{Ioffe, S.} \& \bibinfo{author}{Szegedy, C.}
\newblock \bibinfo{title}{Batch normalization: Accelerating deep network
  training by reducing internal covariate shift}.
\newblock \emph{\bibinfo{journal}{arXiv preprint arXiv:1502.03167}}
  (\bibinfo{year}{2015}).

\bibitem{kaiming_initialization_2015}
\bibinfo{author}{He, K.}, \bibinfo{author}{Zhang, X.}, \bibinfo{author}{Ren,
  S.} \& \bibinfo{author}{Sun, J.}
\newblock \bibinfo{title}{Delving deep into rectifiers: Surpassing human-level
  performance on imagenet classification}.
\newblock In \emph{\bibinfo{booktitle}{Proceedings of the IEEE international
  conference on computer vision}}, \bibinfo{pages}{1026--1034}
  (\bibinfo{year}{2015}).

\bibitem{smith2017cyclical}
\bibinfo{author}{Smith, L.~N.}
\newblock \bibinfo{title}{Cyclical learning rates for training neural
  networks}.
\newblock In \emph{\bibinfo{booktitle}{2017 IEEE winter conference on
  applications of computer vision (WACV)}}, \bibinfo{pages}{464--472}
  (\bibinfo{organization}{IEEE}, \bibinfo{year}{2017}).

\bibitem{jaccard1912}
\bibinfo{author}{Jaccard, P.}
\newblock \bibinfo{title}{The distribution of the flora in the alpine zone 1}.
\newblock \emph{\bibinfo{journal}{New Phytologist}}
  \textbf{\bibinfo{volume}{11}}, \bibinfo{pages}{37--50}
  (\bibinfo{year}{1912}).
\newblock
  \urlprefix\url{https://nph.onlinelibrary.wiley.com/doi/abs/10.1111/j.1469-8137.1912.tb05611.x}.
\newblock
  \eprint{https://nph.onlinelibrary.wiley.com/doi/pdf/10.1111/j.1469-8137.1912.tb05611.x}.

\end{thebibliography}
\end{document}